%% file: IROS2018_Centauro_arxiv.tex
\documentclass[letterpaper, 10 pt, conference]{ieeeconf}  

\IEEEoverridecommandlockouts                              

\makeatletter
\let\NAT@parse\undefined
\makeatother

\usepackage[comma,numbers, compress]{natbib}

\usepackage[hidelinks]{hyperref}
\usepackage{url}
\usepackage{blindtext}
\usepackage{graphicx}
\usepackage[usenames,dvipsnames]{xcolor}
\usepackage[draft]{fixme}
\usepackage{soul}
\fxsetup{theme=color}

\usepackage{amsmath}
\usepackage{amssymb}
\usepackage{bm}

\usepackage{booktabs}
\usepackage{textcomp}
\usepackage{threeparttable}

\usepackage[capitalize]{cleveref}

\usepackage{tikz}
\usetikzlibrary{arrows}
\usetikzlibrary{positioning,calc}
\usetikzlibrary{decorations.pathreplacing}
\usetikzlibrary{decorations.markings}
\usetikzlibrary{fit}
\usetikzlibrary{shapes.callouts}
\usetikzlibrary{shapes.geometric}
\usetikzlibrary{matrix}
\usepackage[per=frac,binary-units=true]{siunitx}

\usepackage[firstpage=true]{background}

\newcommand\copyrighttext{%
        \parbox{\textwidth}{
                \footnotesize
                In Proceedings of IEEE/RSJ International Conference on Intelligent Robots and Systems (IROS), Madrid, Spain, October 2018,\\DOI:  10.1109/IROS.2018.8594509
        }
}

\SetBgContents{\copyrighttext}
\SetBgScale{1}
\SetBgColor{black}
\SetBgAngle{0}
\SetBgOpacity{1}
\SetBgPosition{current page.north}
\SetBgVshift{-0.8cm}

\usepackage[bottom]{footmisc}

\graphicspath{{figures/}}
\newcommand{\etal}{et~al.}
\newcommand{\ie}{i.e.,\ }
\newcommand{\eg}{e.g.,\ }

\usepackage{multirow}

\title{\LARGE \bf
Supervised Autonomous Locomotion and Manipulation\\for Disaster Response with a Centaur-like Robot
}

\author{Tobias Klamt, Diego Rodriguez, Max Schwarz, Christian Lenz, Dmytro Pavlichenko,\\David Droeschel, and Sven Behnke
\thanks{All authors are with the Autonomous Intelligent Systems group of University of Bonn, Germany
        {\tt\small klamt@ais.uni-bonn.de}. This work was supported by the European Union's Horizon 2020 Programme under 
        Grant Agreement 644839 (CENTAURO).}%
}

\begin{document}

\maketitle
\thispagestyle{empty}
\pagestyle{empty}

\begin{abstract}

Mobile manipulation tasks are one of the key challenges in the field of search and rescue (SAR) robotics requiring robots with flexible locomotion and manipulation abilities. Since the tasks are mostly unknown in advance, the robot has to adapt to a wide variety of terrains and workspaces during a mission.
The centaur-like robot Centauro has a hybrid legged-wheeled base and an anthropomorphic upper body to carry out complex tasks in environments too dangerous for humans. 
Due to its high number of degrees of freedom, controlling the robot with direct teleoperation approaches is challenging and exhausting. Supervised autonomy approaches are promising to increase quality and speed of control while keeping the flexibility to solve unknown tasks. We developed a set of operator assistance functionalities with different levels of autonomy to control the robot for challenging locomotion and manipulation tasks. The integrated system was evaluated in disaster response scenarios and showed promising performance.

\end{abstract}


\section{Introduction}

In many SAR scenarios, humans cannot work due to risks such as radiation or collapsing structures. Mobile manipulation robots are promising to help solving tasks in these cases. Respective environments, e.g., the damaged nuclear plant in Fukushima, are mostly man-made but cluttered with debris and unpredictable. Hence, a suitable platform needs to provide a wide range of capabilities to solve occurring tasks and address unforeseen difficulties.

\begin{figure}
	\centering
	\includegraphics[height=0.46\linewidth]{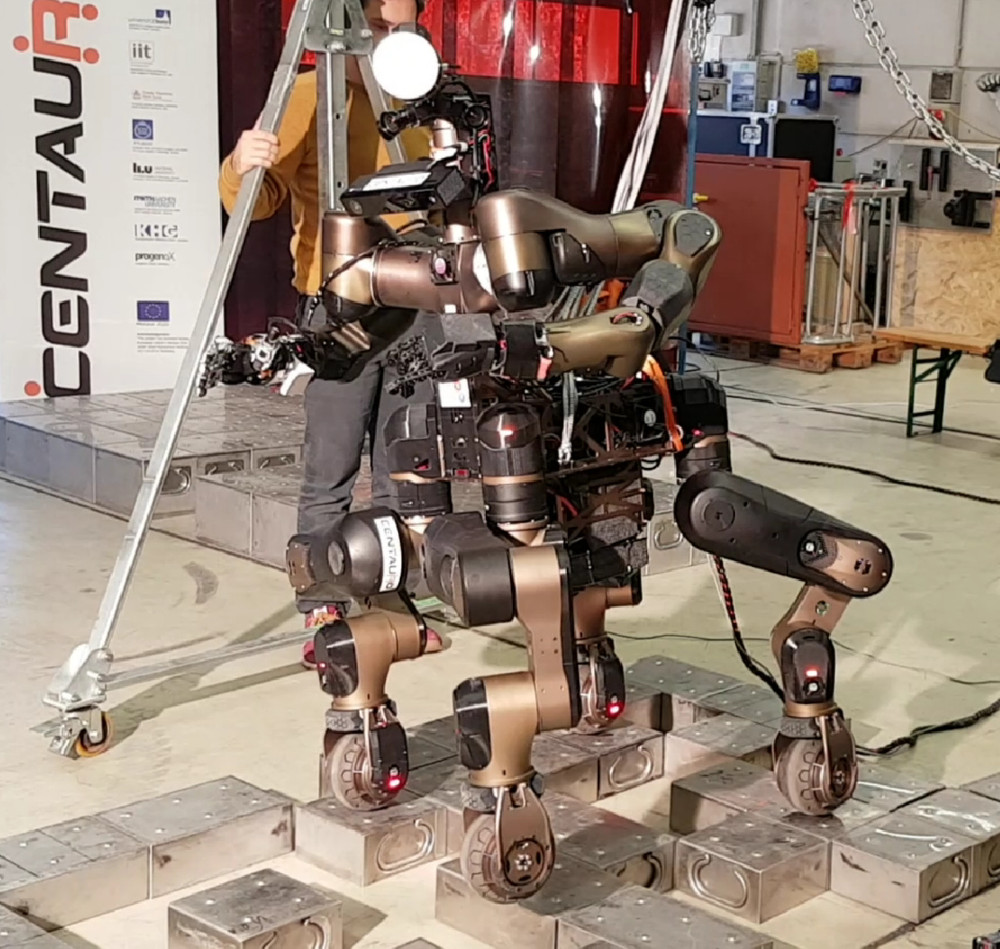}\hspace{0.5ex}
		\includegraphics[height=0.46\linewidth]{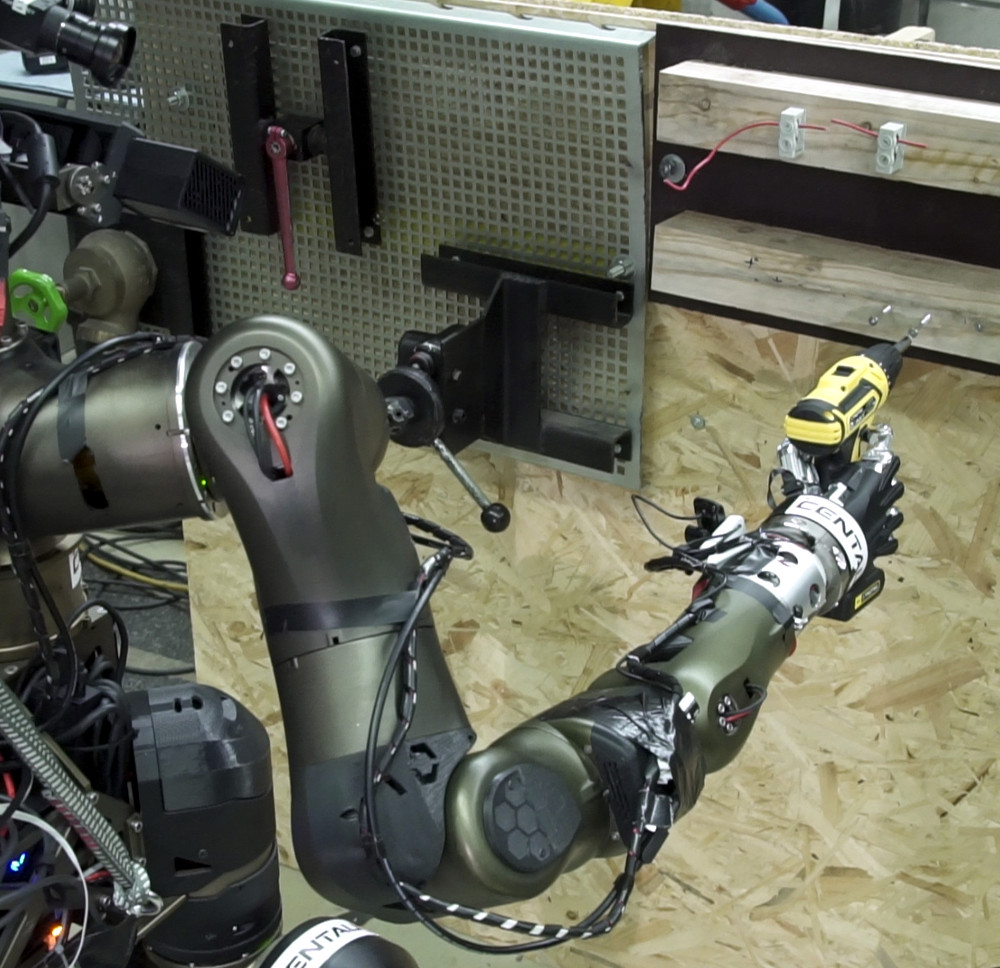}\vspace*{-1.3ex}
	\caption{Centauro robot controlled by our proposed teleoperation system: overcoming a step field (l.) and operating an off-the-shelf power drill (r.).}
	\label{fig:centauro_teaser}
	\vspace{-0.55cm}
\end{figure}

The Centauro robot has been developed in the European H2020 project CENTAURO\footnote{~\url{www.centauro-project.eu}} for such scenarios (\cref{fig:centauro_teaser}). 
Its lower body consists of four articulated legs ending in steerable wheels which allows for omnidirectional driving as well as for stepping locomotion. The anthropomorphic upper body possesses two 7\,DoF arms ending in two hands with different capabilities. One of them is an anthropomorphic Schunk hand which allows for precise manipulation in man-made workspaces. Additionally, several sensors such as a 3D laser scanner and cameras perceive the environment and enable the operators to obtain situation awareness.

Teleoperation of such highly flexible robots is challenging, though. Common approaches, like the control in joint space or Cartesian end-effector space, are only suitable for simple tasks. For more complex tasks, the high number of DoF and typical constraints of multi-legged robots, such as stability and collision avoidance, put a high cognitive load on the operator which may result in slow and dangerous operations. One way to address this challenge is by means of immersive exoskeletons, such as the recently introduced Master Maneuvering System for the Toyota humanoid robot T-HR3\footnote{\url{https://newsroom.toyota.co.jp/en/download/20110424}}. 
These direct control interfaces are as complex as the controlled robot and require a low-latency, high-bandwidth data connection.
Other teleoperation approaches utilize predefined motion primitives. They reduce the operator's cognitive load but the generation of these primitives requires knowledge about specific tasks in advance. Obviously, this restricts the platform flexibility and applicability to unknown tasks. Supervised autonomy is promising to provide fast and reliable control while keeping a high flexibility.

We developed a set of teleoperation interfaces with different levels of autonomy for solving a wide variety of locomotion and manipulation tasks with Centauro. For example, we perform autonomous grasping of unknown tools or semi-autonomous stepping over irregular terrain. Other interfaces with less autonomy include wrist control via a 6D input device. All interfaces provide a high degree of intuition which leads to a limited cognitive load for the operator. This results in less operator failures and extended operation times before the operator needs to recover or must be exchanged.

The integrated functionalities were evaluated in experiments which are typical for disaster-response missions.  Locomotion capabilities were evaluated in tasks like driving up a ramp, overcoming a gap, and moving through an irregular step field. Manipulation interfaces were evaluated in experiments like grasping and using different power tools, physically connecting and disconnecting objects such as electrical plugs, or scanning surfaces, \eg for radiation. A combination of locomotion and manipulation capabilities was required for opening and passing a door. Most of the tasks were solved quickly and without previous training.


\section{Related Work}

Several mobile manipulation robots have been developed for SAR missions. Those robots vary in their locomotion strategy as well as their manipulation setup. Pure wheeled, tracked or legged robots can either overcome long, sufficiently even distances quickly or can navigate in challenging terrain with isolated footholds, but a combination of both capabilities is only available for hybrid driving-stepping locomotion platforms. Manipulation capabilities depend on the number and design of arms and especially the end-effectors. However, independent from robot design details, key to the applicability in a wide range of scenarios are teleoperation interfaces which ideally enable teleoperators to use all robot capabilities while keeping the cognitive load low and the applicability to unknown tasks high. In 2015, the DARPA Robotics Challenge (DRC) pushed research teams to develop robots that are capable of performing several mobile manipulation tasks which indirectly put the focus on the development of suitable teleoperation interfaces.

RoboSimian~\cite{hebert2015mobile} is a quadrupedal robot with four generalized limbs, developed for the DRC. Each limb ends in an under-actuated hand which allows for solving both stepping locomotion and manipulation tasks. Furthermore, RoboSimian has two active wheels at its trunk and two caster wheels at its limbs which allow for driving on even terrain. The operator interface is a standard laptop from which the operator can design, parametrize, and sequence predefined behaviors. The DRC winner robot DRC-HUBO~\cite{zucker2015general} and the third placed platform CHIMP~\cite{stentz2015chimp} both have roughly anthropomorphic bodies. Both are capable of walking and driving via additional wheels/tracks on their body. Both robots have two arms which end in hands with three fingers. Operation of DRC-HUBO is apportioned among three operators with different tasks which control the robot by selecting and adapting predefined poses while CHIMP is operated through task-specific motions which are configured through wizards by the operator before their execution. 

Our centaur-shaped robot Momaro~\cite{schwarz2017nimbro} came in 4th in the DRC using multiple teleoperation interfaces and showed autonomy solving known tasks at the DLR SpaceBot Cup~\cite{schwarz2016supervised}. Similar to Centauro, it has four legs ending in steerable wheels and an anthropomorphic upper body. In contrast to Centauro, it lacks hip yaw joints for the legs which restricts stepping capabilities. Moreover, its two arms end in 4-finger grippers which cannot provide the grasping capabilities of a human hand. Driving locomotion can be controlled by a joystick; leg motions are predesigned or can be controlled during mission in joint space or Cartesian end-effector space. Furthermore, a semi-autonomous stepping controller was presented which relies on perceived terrain heights~\cite{schwarz2016hybrid}. For telemanipulation, the operator used two hand-held controllers with magnetic trackers whose movements were projected to the robot arms. Although this approach appears to be intuitive, the operator experienced a high cognitive load due to imprecision in the motion mapping and the lack of feedback. Grasping was controlled by predefined gripper configurations.

Momaro can be seen as the predecessor of Centauro. Key design features, such as the general lower and upper body kinematics and the sensor setup, were transferred. 
Weak points, such as chosen actuators, missing hip yaw joints or restricted end-effectors, were improved. 

Regarding the operator interface, we enrich Momaro's teleoperation interfaces by adding more intuitive control devices.
Additionally, we focus on solving unknown tasks by incorporating flexible autonomous capabilities. An overview over robot control approaches with different levels of autonomy is given by Kiu \etal~\cite{liu2013robotic}.
As shown recently by Marturi \etal~\cite{marturi2016} and earlier by Leeper \etal~\cite{leeper2012}, the task efficiency and accuracy are improved by incorporating further interfaced and autonomous functionalities. 
Muelling \etal~\cite{muelling2017}, for example, developed an integrated system of computer vision with manipulation capabilities, 
in which known objects with simple geometries are recognized, localized and grasped using depth images.
Peer \etal~\cite{peer2008multi} and Salvietti \etal~\cite{salvietti2017multicontact} present telemanipulation approaches by mapping operator hand configurations to the robot hand and provide force feedback.
Although such interfaces seem intuitive, they generally require a large amount of operator training to provide a satisfying grasp quality.
Havoutis \etal~\cite{havoutis2016} learn manipulation tasks online for semi-autonomous teleoperation applications where large communication latency make direct teleoperation unfeasible.
In our current system, we use both perception and learning approaches to enrich our teleoperation capabilities.


\section{Hardware}
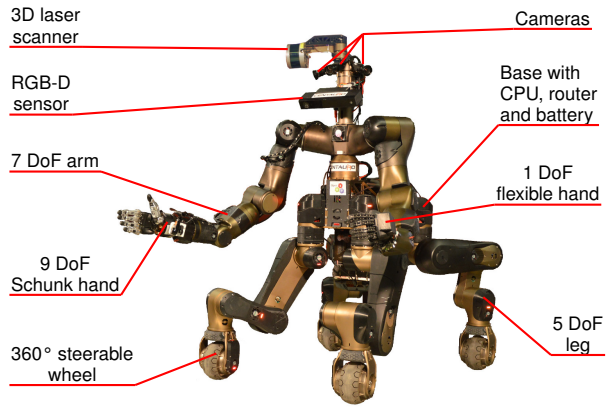
\begin{figure}
\centering
\scalebox{1.0}{\input{figures/figures_section1/centauro_overview.pgf}}
\caption{The Centauro robot.}
\vspace{-0.5cm}
\label{fig:centauro_robot}
\end{figure}

\begin{figure*}[ht]
	\centering
	\scalebox{1.15}{\input{figures/teleoperation_architecture/situation_awareness.pgf}}
	\vspace{-0.25cm}
	\caption{Environment and robot state visualization for the operators.}
	\label{fig:situation_awareness}
	\vspace{-0.4cm}
\end{figure*}
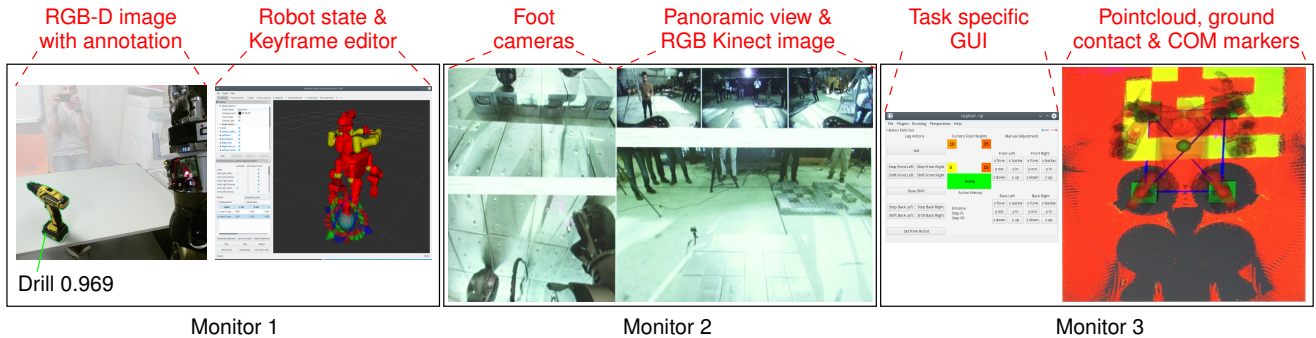

Centauro (\cref{fig:centauro_robot}) was designed by the Istituto Italiano di Tecnologia (IIT), bringing together Momaro's kinematic concept and Walk-Man's compliant actuation~\cite{TsagarakisWALKMANhighperformance2017}. Centauro's kinematic designs aims to provide a wide range of locomotion and manipulation capabilities to solve any occurring disaster response task while the robot size is suitable for man-made environments and workspaces~\cite{BaccelliereDevelopmenthumansize2017}.

Centauro's lower body features four articulated 5-DoF legs which end in 360\textdegree\,steerable, directly driven wheels. This design allows for both omnidirectional driving and stepping locomotion. In addition, Centauro can perform locomotion actions which are not possible for driving or legged platforms, such as shifting individual feet while keeping ground contact. 

The anthropomorphic upper body consists of a torso yaw joint and two 7-DoF arms ending in end-effectors with different capabilities. The right end-effector is a 9-DoF anthropomorphic Schunk hand which allows for dexterous, human-like manipulation~\cite{ruehl2014experimental}. The left end-effector is a flexible 1-DoF  SoftHand which can be used for robust manipulation~\cite{catalano2014adaptive}. The overall upper body design results in a workspace equal to an adult-sized human.

Centauro's head comprises a \emph{Microsoft Kinect V2} RGB-D sensor~\cite{fankhauser2015}, an array of three \emph{PointGrey BlackFly BFLY-U3-23S6C} wide-angle color cameras, and a rotating \emph{Velodyne Puck VLP-16} 3D laser scanner with a spherical field-of-view. A \emph{VectorNav VN-100} IMU is mounted in the torso. Two further RGB cameras were mounted under the robot base to obtain a view on the feet. Furthermore, the robot base incorporates three computing units as well as the communication routers and the robot battery.


\section{Teleoperation Architecture}

Although the considered disaster environments are too dangerous for a human to work in, the human capabilities of situation assessment, mission planning, and his experience are key to a successful SAR mission. The teleoperation interface enables the operators to transfer these capabilities into the scene by providing them an awareness of the situation and enabling them to control the robot. Both require a communication infrastructure, since a direct line of sight is not available.

\subsection{Communication}

For data transmission between the operator station and the robot, we use Ethernet connection or a standard IEEE 802.11ac 5\,GHz WiFi link.
All communication takes place using ROS, which is either directly accessed using ROS network transparency,
or encoded with FEC for robustness using the \texttt{nimbro\_network} developed for Momaro~\cite{schwarz2017nimbro}.
For extending the reach, a WiFi repeater can be carried by the Centauro robot and dropped at
an appropriate location.

\subsection{Situation Awareness}

We developed several visualizations of the environment and the robot state to provide good situation awareness for the operators. RGB camera images from the three cameras in the robot head are arranged to show a panoramic view from the robot head perspective which is helpful for a general scene understanding. 
In addition, images from the two RGB cameras under the robot base are arranged to give a detailed assessment for the terrain under the robot base which was key to a safe stepping locomotion operation. We rotated the image of the camera showing the two rear feet by 180\textdegree\,for intuitive visualization (\cref{fig:situation_awareness}). 
Moreover, laser scanner measurements are processed to registered point clouds which are visualized in RVIZ (Sec.\,\ref{sec:3d_laser_scan_assembly}).
This visualization is helpful for both locomotion and manipulation tasks. Finally, colored RGB-D point clouds are displayed to support manipulation. Those are enriched by semantics from the object detection (Sec.\,\ref{sec:object_segmentation}). 

The robot state is visualized by applying measured joint angles and IMU data to a 3D robot model in RVIZ. Further information, such as foot ground contact detection and the robot center of mass (CoM) are also displayed. We developed multiple robot control GUIs for different task classes. All visualization elements were arranged on three monitors as shown in~\cref{fig:situation_awareness}.

\subsection{Control Interfaces}

We propose multiple locomotion and manipulation control interfaces which are suitable for different task classes. The whole set of control interfaces aims at enabling the operator to solve as many---previously known and unknown---typical disaster response tasks as possible. Hence, a key requirement is to address the whole range of kinematic capabilities of the robot while keeping the control itself intuitive. Different levels of autonomy are utilized to fulfill these requirements. The individual control interfaces are presented in Sec.\,\ref{sec:locomotion_control} and Sec.\,\ref{sec:manipulation_control}. Some of them require processed sensor input which is described in the following.


\section{Advanced Environment Perception}
\label{sec:environment_perception}

The chosen sensor setup produces data of several types. While some sensor measurements, such as foot camera images, can be directly shown to the operators, other data is processed. The results serve as more intuitive visualizations or as input for some of the autonomous control functions.

\subsection{Ground Contact Detection}

To understand the robot positioning in challenging terrain and to enable semi-autonomous stepping, it is helpful to detect, if a foot has ground contact. By measuring the joint torques of the respective leg and by applying a forward dynamics approach, we compute the 6D force vector which is applied to the foot. The vertical force component is extracted and compensated for gravity. If the resulting force exceeds a given threshold, ground contact is detected.

\subsection{Laser-based 3D Mapping and Localization}
\label{sec:3d_laser_scan_assembly}

Laser range measurements from the 3D rotating laser scanner are aggregated to a dense 3D map of the environment using our local multiresolution surfel grid approach~\citep{Droeschel2017104}. The laser provides \texttildelow300,000 range measurements per second with a maximum range of 100\,m and is rotated at 0.1 rotations per second, resulting in a dense omnidirectional 3D scan per halve rotation. We acquire one full 3D scan every five seconds and compensate for sensor motion during acquisition by incorporating IMU measurements. 

\begin{figure}
 \centering
 \setlength{\tabcolsep}{0pt}
 \includegraphics[height=0.42\linewidth]{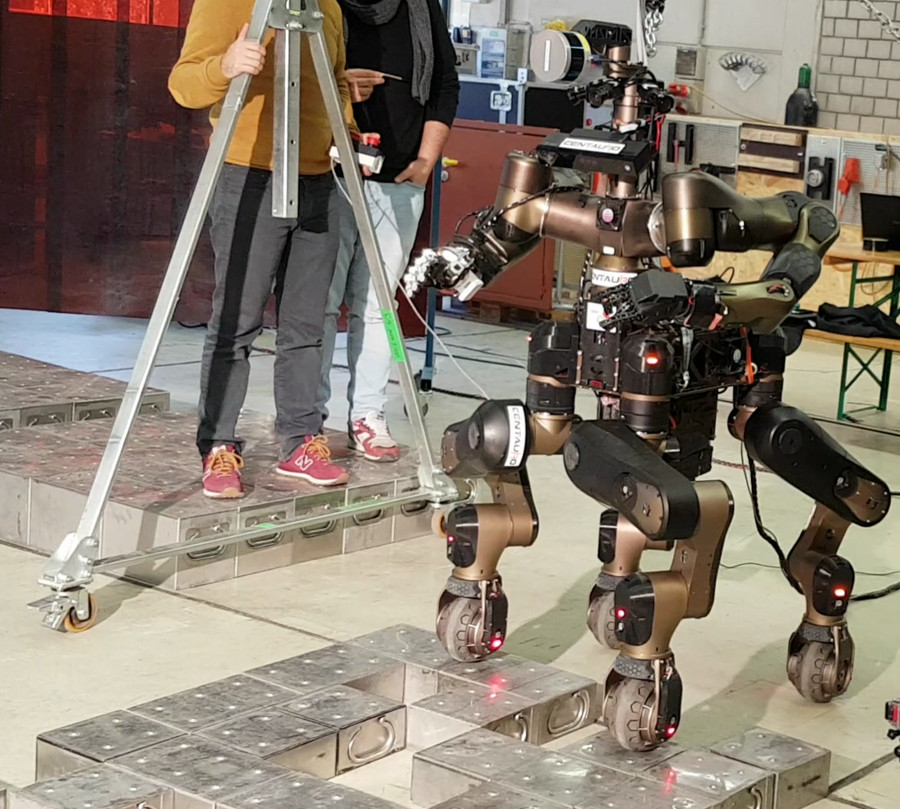}
 \includegraphics[height=0.42\linewidth]{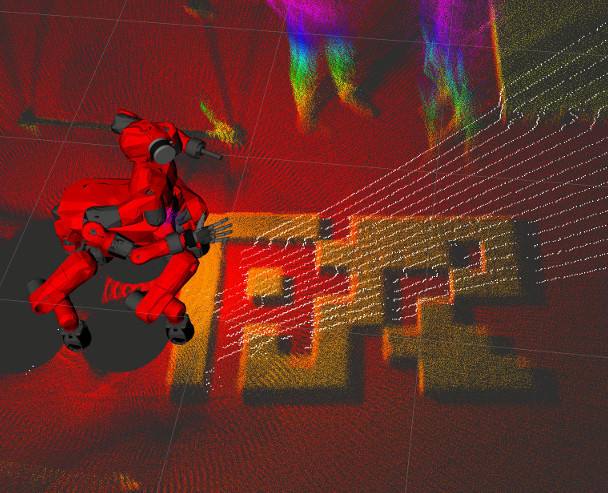}\vspace*{-1ex}
 \caption{Centauro robot traversing a step field. Left: photo of the scene, right: laser-based 3D map (colored points) and current scan (white points).}
 \vspace{-0.5cm}
 \label{fig:perception:laser}
\end{figure}

Consecutive scans are registered to a dense egocentric map. The resulting egocentric maps from different view poses form nodes in a pose graph to allow for allocentric mapping of the environment. They are connected by edges representing spatial constraints, which result from aligning these maps with each other. The global registration error is minimized using graph optimization. The resulting 3D map allows for localizing the robot in an allocentric frame. A resulting 3D map is shown in~\cref{fig:perception:laser}.

\subsection{Object Segmentation}
\label{sec:object_segmentation}
We apply our object segmentation approach to RGB images from the \emph{Kinect V2}~\citep{schwarz2018fast}. This approach is able to produce pixel- (or point-)wise segmentation directly.
It uses the RefineNet~\cite{lin2016refinenet} architecture, 
which addresses the problem of low spatial resolution in later stages of the CNNs by
subsequently upsampling and merging higher-level feature maps with lower-level
features of higher spatial resolution---creating a representation of the input
image with both highly semantic information and high spatial resolution,
which is well suited for semantic segmentation.

\begin{figure}
 \centering
 \setlength{\tabcolsep}{0pt}
 \includegraphics[width=0.49\linewidth]{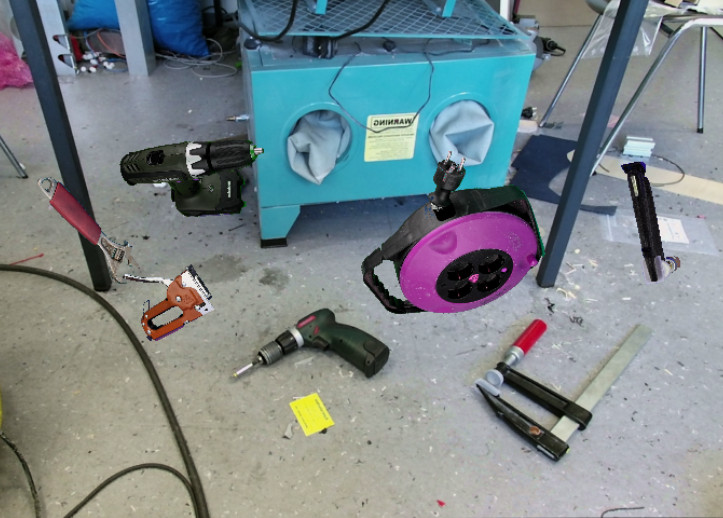}
 \includegraphics[width=0.49\linewidth]{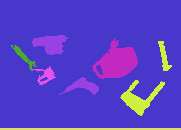}\vspace*{-1ex}
 \caption{Scene synthesis. 
 Synthetic training scene generated by inserting new objects into the scene.
 The right image shows the resulting color image, the left one shows
  synthetic ground truth for training the segmentation model.}
 \label{fig:perception:synthesis}
\end{figure}

We address the lack of large amounts of training data by generating new training scenes using data captured from a turntable
setup. Automatically extracted object segments are inserted into precaptured
scenes (\cref{fig:perception:synthesis}).

\subsection{Pose Estimation}
\label{sec:perception:posenet}
\begin{figure}
  \raisebox{-0.5\height}{\includegraphics[width=.48\textwidth]{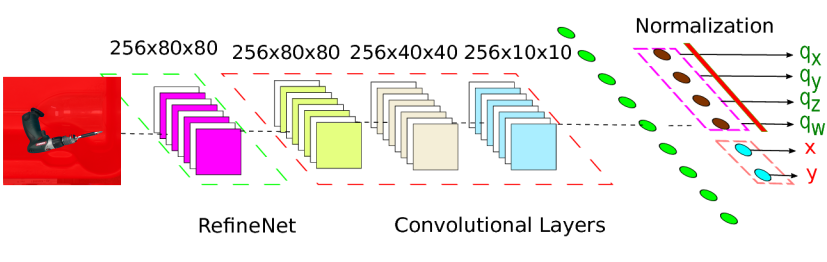}}\vspace*{-1ex}\hfill
  \caption{Pose estimation network architecture.
  }
  \vspace{-0.5cm}
  \label{fig:perception:posenet_architecture}
\end{figure}

For predicting poses efficiently, we augment the semantic segmentation pipeline with an
additional CNN to estimate the object 5D pose (rotational, and X and Y of the translational components) from RGB-D crops of the objects from the scene. 
Those crops are extracted from the bounding boxes of detected contours.
To encode the segmentation results, pixels classified as non-object are pushed towards red (Fig.~\ref{fig:perception:posenet_architecture}).
This representation allows the network to focus on the specific object for which the pose should be estimated.
The pretrained RefineNet network from the semantic segmentation is used to extract features.
To generate the ground truth poses for training the network, the data acquisition pipeline described in~\cite{schwarz2018fast} was extended to record turntable poses automatically and fuse captures with different object poses or different objects with minimal user intervention.


\section{Locomotion Control}
\label{sec:locomotion_control}

Centauro's lower body design allows for omnidirectional driving as well as stepping locomotion and, hence, provides a wide range of locomotion capabilities which have to be addressed by the respective control interface. Driving locomotion allows for fast, energy efficient and stable navigation on sufficiently even terrain while stepping locomotion increases the platform's capabilities to terrains where only isolated footholds are available. Besides the listed control interfaces, we developed a hybrid driving-stepping locomotion planner~\cite{klamt2017anytime, klamt2018planning} which lifts the level of autonomy even higher but has not been evaluated on the real platform, yet.

\subsection{4D Joystick}

\begin{figure}
 \centering
 \includegraphics[width=0.30\linewidth]{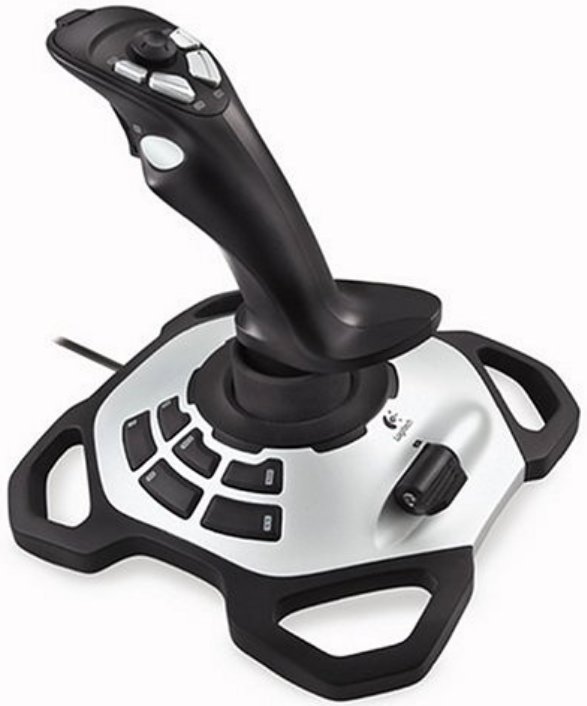}
  \includegraphics[width=0.23\linewidth]{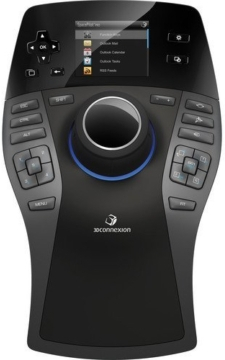}
  \includegraphics[width=0.38\linewidth]{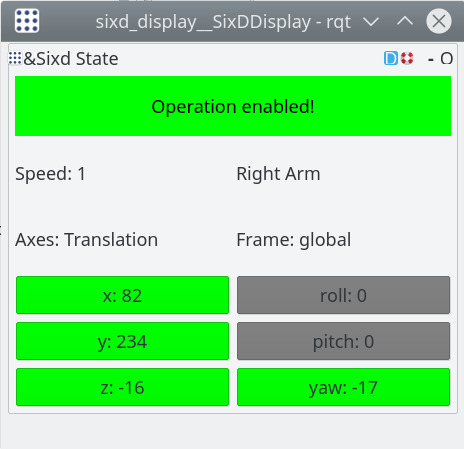}\vspace*{-1ex}
 \caption{Operator input devices. Left: \emph{Logitech Extreme 3D Pro} for omnidirectional driving control, center and right: \emph{3DConnexion SpacePilot Pro} and respective operator GUI for dexterous manipulation.}
 \vspace{-0.5cm}
 \label{fig:joystick_6D_mouse}
\end{figure}

Omnidirectional driving can be controlled by a \emph{Logitech Extreme 3D Pro} joystick with four axis (\cref{fig:joystick_6D_mouse}). Robot base velocity components $v_x$, $v_y$ and $v_\theta$ are mapped to the three corresponding joystick axis. Foot-specific velocities and orientations are derived from this robot base velocity and the individual foot positions. The joystick throttle controller jointly scales all three velocity components.

\subsection{Keyframe Editor}
\label{sec:keyframe_editor}

A keyframe editor generates robot motions by interpolating between given keyframes~\cite{schwarz2016supervised}.
Keyframes for joint groups (\eg the front left leg) can either be specified in joint space or in Cartesian end-effector space. Longer motion sequences can be designed by queuing keyframes. The operators can either predefine keyframes, modify them during the mission and send them to the robot, or modify the robot configuration live. The RVIZ-based GUI (\cref{fig:locomotion_control}) allows for keyframe definition by either graphically moving joint group markers with the mouse or by entering numerical values for desired joint angles or end-effector positions.

\subsection{Semi-autonomous Stepping Locomotion}
\label{sec:semi_autonomous_stepping}

Stepping locomotion can be controlled by a semi-autonomous controller. It provides a set of motions which can be triggered by the operator. Available motions are: step with a chosen foot, drive a chosen foot forward, and shift the robot base forward. For stepping motions, the controller balances the robot by shifting the robot base longitudinally and laterally, and by rolling it around its longitudinal axis. If a stable pose is established, the stepping foot is lifted, extended by a given length and lowered. The lowering motion stops when ground contact is detected. Hence, the robot adapts to the terrain automatically. The proposed controller triggers queues of keyframes, as described in Sec.\,\ref{sec:keyframe_editor}. 

\begin{figure}
	\centering
	\input{figures/locomotion_control/locomotion_control.pgf}\vspace*{-1ex}
	\caption{Left: Keyframe editor, right: semi-autonomous stepping GUI.}
	\label{fig:locomotion_control}
	\vspace{-0.5cm}
\end{figure}
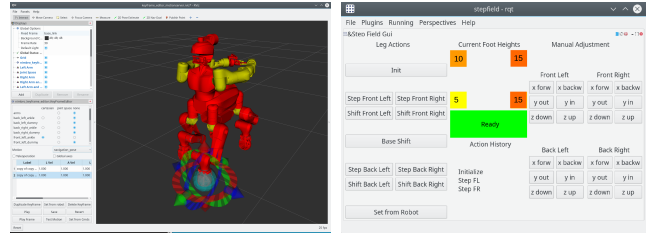

We developed an intuitive GUI which provides buttons to trigger the described motions (\cref{fig:locomotion_control}). 
It also contains buttons to manually move individual feet in Cartesian space. Moreover, the GUI displays detected terrain heights under the four feet and a history of the recently triggered motions which is helpful to execute repetitive motion sequences.

\subsection{Motion Execution}

The Centauro robot uses a keyframe interpolation method developed for Momaro to generate joint space trajectories~\cite{schwarz2017nimbro}. Keyframes consist of joint space or 6D Euclidean space poses for each of the robot's limbs. The interpolation system produces jerk-free joint-space trajectories obeying velocity and acceleration constraints set per keyframe.


\section{Manipulation Control}
\label{sec:manipulation_control}

Regarding the robot's manipulation capabilities, the Centauro system possesses several levels of autonomy: 
starting at low-level direct joint control; 
over inverse kinematics control with end-effector poses coming from either an 6D input device, or 6D markers on the screen; 
keyframe motions with collision avoidance; 
and finally, autonomous pick-and-place actions triggered by the operator.
For manipulation, we also use the same interface as described in Sec.\,\ref{sec:keyframe_editor}.
Thus, we will only describe in this section the novel 6D input interface and the autonomous grasping capabilities.

\subsection{Dexterous Wrist Manipulation}

We developed a user interface for dexterous manipulation using a \emph{3DConnexion SpacePilot Pro} which is a 6D input device with additional buttons (\cref{fig:joystick_6D_mouse}).
The interface establishes the connection between the device and a motion player, which interpolates between the desired and current poses and executes the motion.

The following control parameters can be easily adjusted by the GUI (\cref{fig:joystick_6D_mouse}) or by the device buttons: the \textit{controlled end-effector} (\eg a wrist for arm control or an ankle for leg control), the \textit{reference frame} (e.g., robot base frame, end-effector frame, or a custom frame), the translational and rotational \textit{axes} in which the end-effector is allowed to move, and the maximum end-effector \textit{speed}.

This teleoperation interface is well suited for manipulation tasks where very precise arm movement along certain directions is required (\eg moving the arm along a plane surface or turning an object around a specified axis). 

\subsection{Autonomous Grasping}
\begin{figure}[b]
	\centering
	\input{figures/manipulation/overview.pgf}\vspace*{-1ex}
	\caption{Overview of autonomous manipulation: integrated sensors (red), perception modules (purple), and manipulation planning (yellow).}
	\vspace{-0.5cm}
	\label{fig:manipulation:overview}
\end{figure}
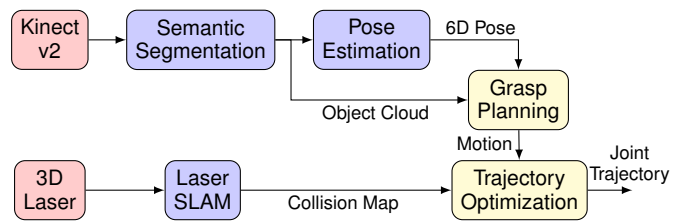

To achieve autonomous manipulation, several components need to be developed and integrated.
We propose a pipeline composed of: \textit{semantic segmentation} (Sec.~\ref{sec:object_segmentation}), \textit{pose estimation} (Sec.~\ref{sec:perception:posenet}), and \textit{grasp planning} that generates a feasible motion (set of keyframes), which later is combined with a \textit{trajectory optimization} that produces the final joint trajectory given a collision map generated by the \textit{laser SLAM} (Sec.~\ref{sec:3d_laser_scan_assembly}) perception module (Fig.~\ref{fig:manipulation:overview}).

\subsubsection{Grasp Planning}
\label{sec:grasp_gen}
Our grasp planning method is based on the observation that objects within a category exhibit several similarities in their extrinsic geometry.
We transfer grasp poses from a known instance---called the canonical model---to novel instances of the same category.

\begin{figure}
 	\centering
 	\includegraphics[width=1.0\linewidth]{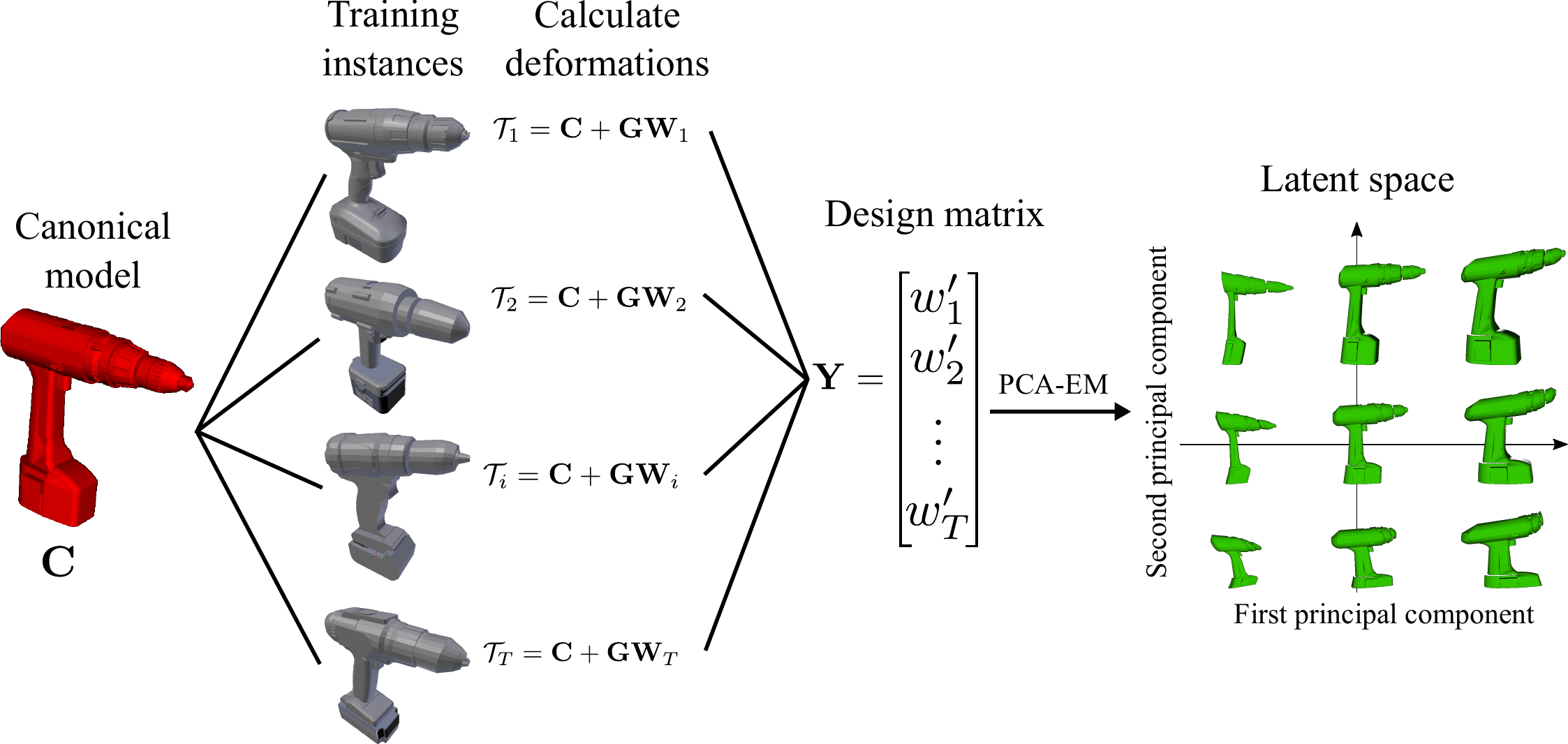}\vspace*{-1ex} 
 	\caption{The latent shape space is built by applying PCA-EM on a matrix containing the deformation fields $\mathbf{w_i}$ of each training instance toward the canonical model.}
 	\vspace{-0.5cm}
 	\label{fig:CLS_training}
 \end{figure}

This transfer happens as the result of a non-rigid registration method based on a learned latent shape space.
For building this latent shape space, we define a single canonical model for the category, 
and calculate the deformation fields relating the canonical model to all other instances by using Coherent Point Drift (CPD)~\cite{myronenko2010point}.
This provides a single matrix whose number of elements equals the number of points in the canonical model for each instance.
A design matrix containing all deformation fields is consequently assembled as column vectors.
Finally, we apply Principle Component Analysis Expectation Maximization (PCA-EM) on the design matrix to find a lower-dimensional manifold of deformation fields, \ie the latent shape space (\cref{fig:CLS_training}).
 
We add a global rigid transformation for each instance to reduce the impact of minor misalignments in the pose between the canonical shape and the observed shape.
For registering a new instance, we use gradient descent to simultaneously optimize for pose and shape.
In general, we aim for an aligned dense deformation field that matches best the canonical model toward the observed instance. 
Associated grasping control poses of the canonical model are also transformed to the observed instance and used for the final grasping motion.
We orthonormalize the transformed poses since the warping process can violate the orthogonality of the orientation. 

\begin{figure}
	\centering
	\scriptsize
	a.)\includegraphics[height=2.cm]{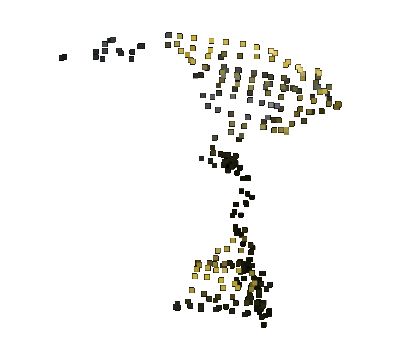}\hfill
	b.)\includegraphics[height=2.cm]{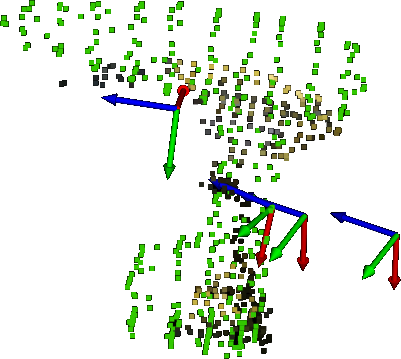}\hfill
	c.)\includegraphics[height=2.cm]{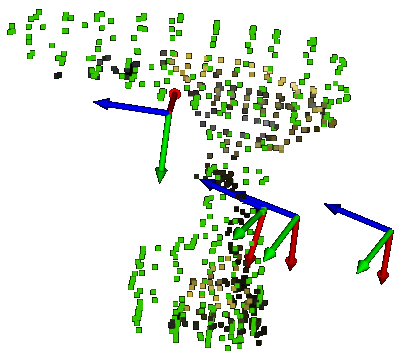}\hfill
	d.)\includegraphics[height=2.cm]{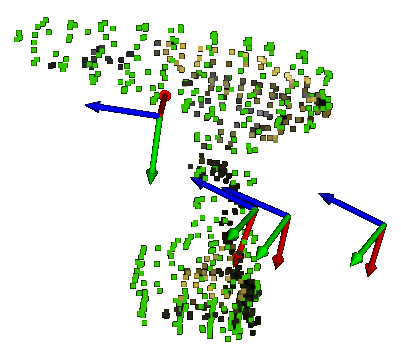}\hfill
	e.)\includegraphics[height=2.cm]{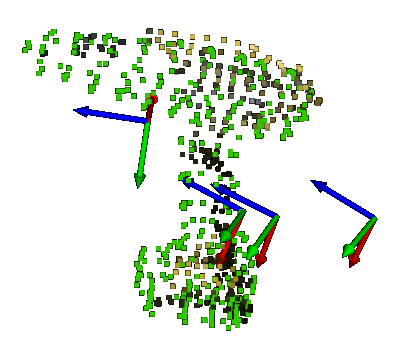}\hfill
	f.)\includegraphics[height=2.cm]{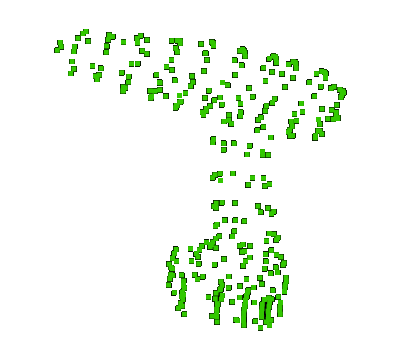}
	\caption{Transferring grasping knowledge to the presented novel drill. a) novel view; b)-e) grasping control poses of the canonical model are transformed;
		f) inferred shape.}.
	\label{fig:eval:CLS}
	\vspace{-0.5cm}
\end{figure}

\cref{fig:eval:CLS} illustrates how the canonical model and associated grasping control poses of a \textit{Drill} category are warped to fit to the observed point cloud. A complete analysis and discussion of this method is available in~\cite{Rodriguez2018} and~\cite{Rodriguez2018b}.

\subsubsection{Trajectory Optimization}
\label{sec:traj_opt}

We use arm trajectory optimization to generate collision free and fast arm trajectories with low actuator load. 
Our approach~\cite{Pavlichenko2017} is based on Stochastic Trajectory Optimization for Motion Planning (STOMP)~\cite{Kalakrishnan2011}.
The method receives a point cloud describing the environment and an initial trajectory as input. It outputs a trajectory, that is optimized with respect to a cost function. The initial trajectory may be very na{\"i}ve, for example a straight interpolation between the start and the goal configuration.
The trajectories are represented as sets of keyframes in joint space.
The optimization is performed in an iterative manner in order to gradually minimize the costs. In contrast to the original STOMP, our cost function is defined as a sum of costs of \textit{transitions} between the consequent keyframes instead of the keyframes themselves. The cost function includes trajectory duration, collision avoidance, and required joint torques. Since cost components are normalized, they can be weighted to introduce a prioritized optimization. 

\begin{figure}[b]
	\centering
	\includegraphics[width=0.43\linewidth]{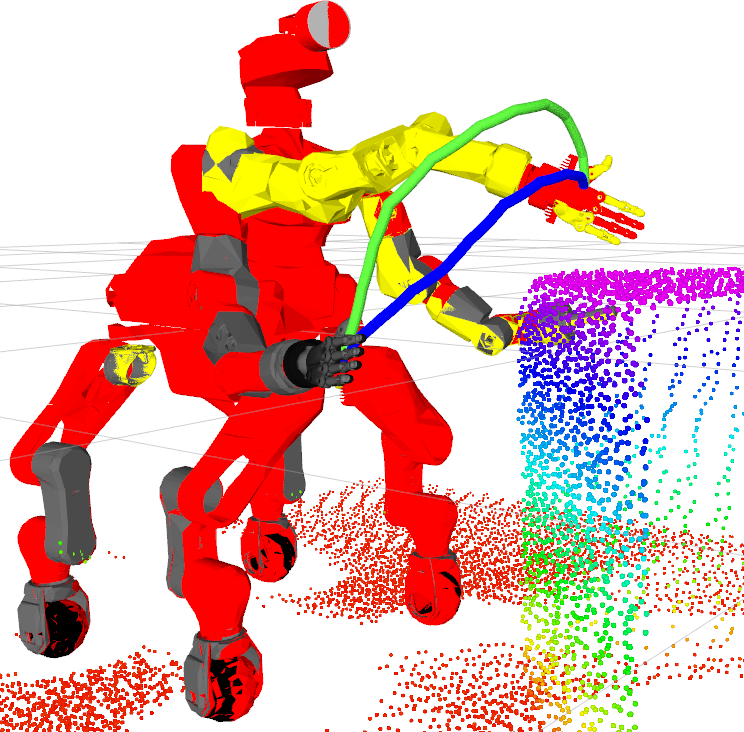} 
	\caption{Two qualitatively different trajectories generated by our trajectory optimization: priority on obstacle avoidance (green) and priority on trajectory duration (blue). The point cloud visualizes the environment.
		 }
		 \vspace{-0.5cm}
	\label{fig:manipulation:arm_trajectories}
\end{figure}
For collision avoidance, we assume the robot base and the environment to be static and describe both with signed Euclidean Distance Transforms (EDT) which allow for fast collision checking against the moving robot parts, represented as spheres. An example is shown in~\cref{fig:manipulation:arm_trajectories}.


\section{Evaluation}
We evaluated the Centauro system with task-level tests at facilities
of the Kerntechnische Hilfsdienst GmbH in Karlsruhe, Germany, which is a provider
of systems and knowledge for disaster response in nuclear power plants.
All tasks were performed without direct visual contact such that the operators had to rely on information provided by our interfaces.
There were no training runs for any of the tasks.
A video with footage from the experiments is available online\footnote{\url{https://www.ais.uni-bonn.de/videos/IROS_2018_Centauro}}.
The results are summarized in \cref{tab:eval_tasks}.

\subsection{Locomotion Tasks}
\begin{figure}[b]
 \centering\newlength{\doorheight}\setlength{\doorheight}{3cm}
 \includegraphics[height=\doorheight, clip, trim=0 0 0 0]{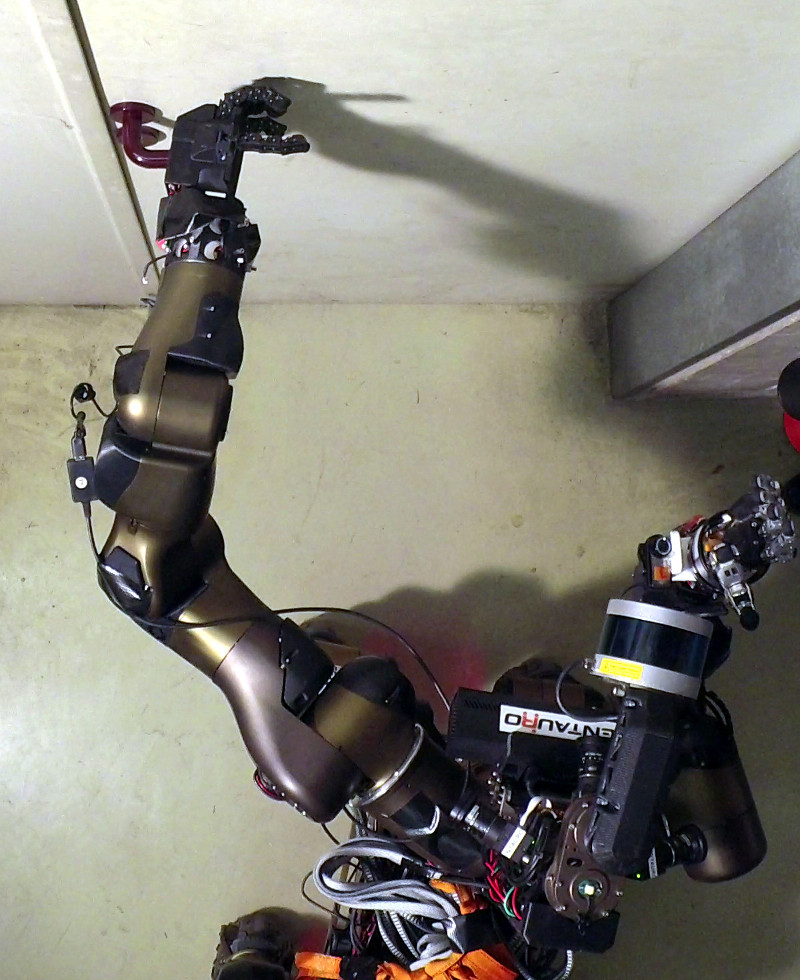}
 \includegraphics[height=\doorheight, clip, trim=0 0 0 0]{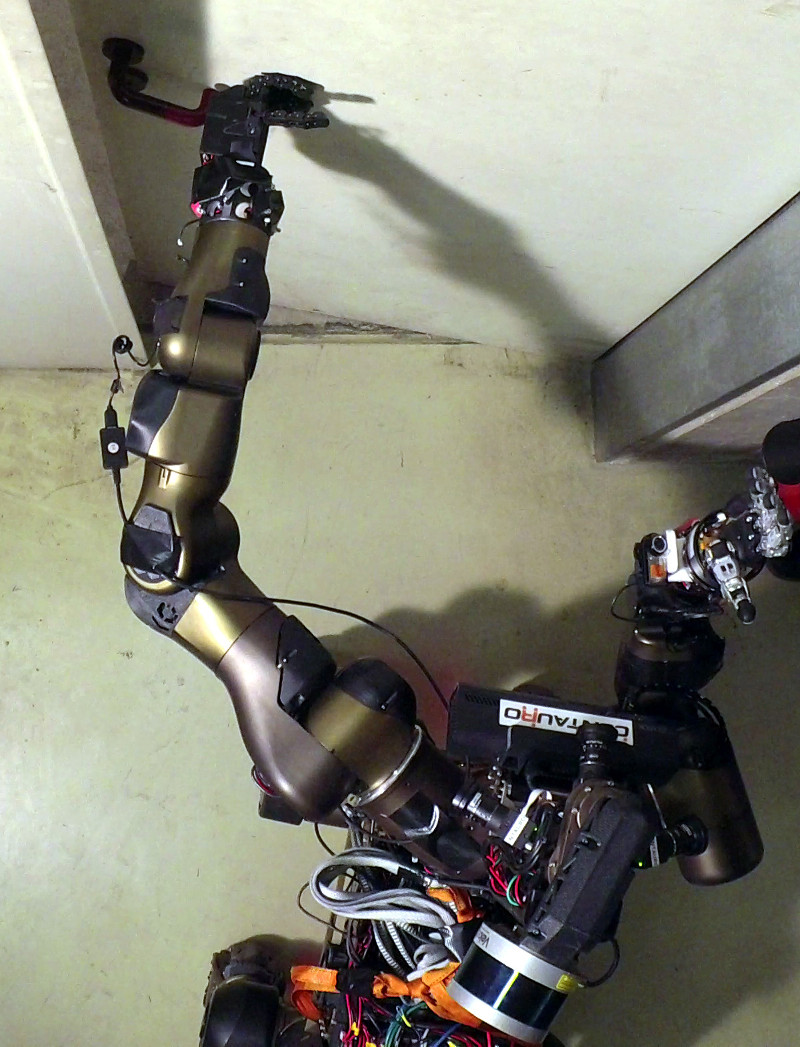}
 \includegraphics[height=\doorheight, clip, trim=0 0 0 0]{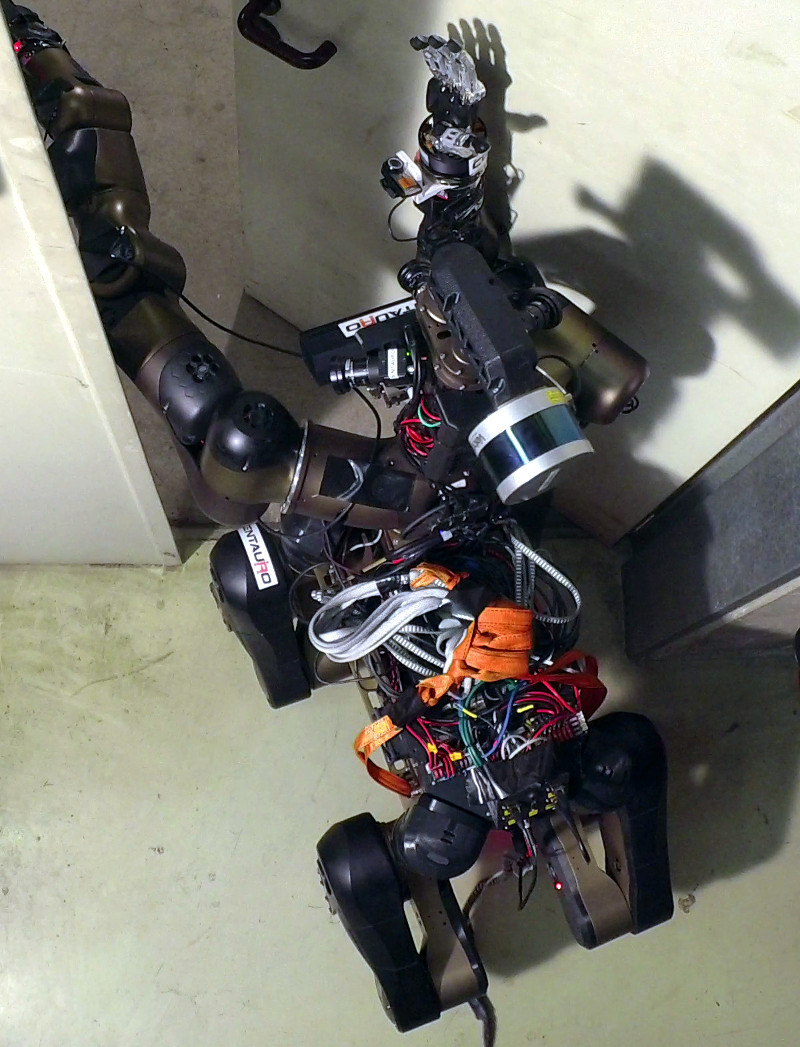}\vspace*{-1ex}
 \caption{Opening the door and driving through it.
 }
 \label{fig:eval_door}
\end{figure}

The tested locomotion tasks mainly focused on proving that the robot can effectively navigate different complex terrain types.
In the simplest task, the robot was required to drive up a ramp with
$20^\circ$ incline, which was accomplished using joystick teleoperation.
In the door experiment (\cref{fig:eval_door}), the robot had to open a door and drive through it.
The manipulation part was accomplished using the 6D mouse control without
any problem.

\begin{table}
 \centering
 \caption{Evaluated tasks.}\label{tab:eval_tasks}
 \begin{tabular}{lclc}
  \toprule
  \multicolumn{2}{c}{Locomotion} & \multicolumn{2}{c}{Manipulation}\\
  \cmidrule(lr){1-2}\cmidrule(lr){3-4}
  Task & Success/Tries & Task & Success/Tries \\
  \cmidrule(lr){1-2}\cmidrule(lr){3-4}
  Door & 3/3 & Surface detection & 2/2 \\
  Ramp & 3/3 & Plug & 2/3 \\
  Gap & 3/4 &  Screw driver & 3/3 \\
  Step field & 2/2 & \multirow{2}{6em}{Autonomous grasping} & \multirow{2}{*}{7/14} \\
  Stairs & 0/1 \\
  \bottomrule
 \end{tabular}
  \vspace{-0.3cm}
\end{table}

\begin{figure}[t]
  \centering\newlength{\gapheight}\setlength{\gapheight}{3.1cm}
  \includegraphics[height=\gapheight,clip,trim=60 0 0 0]{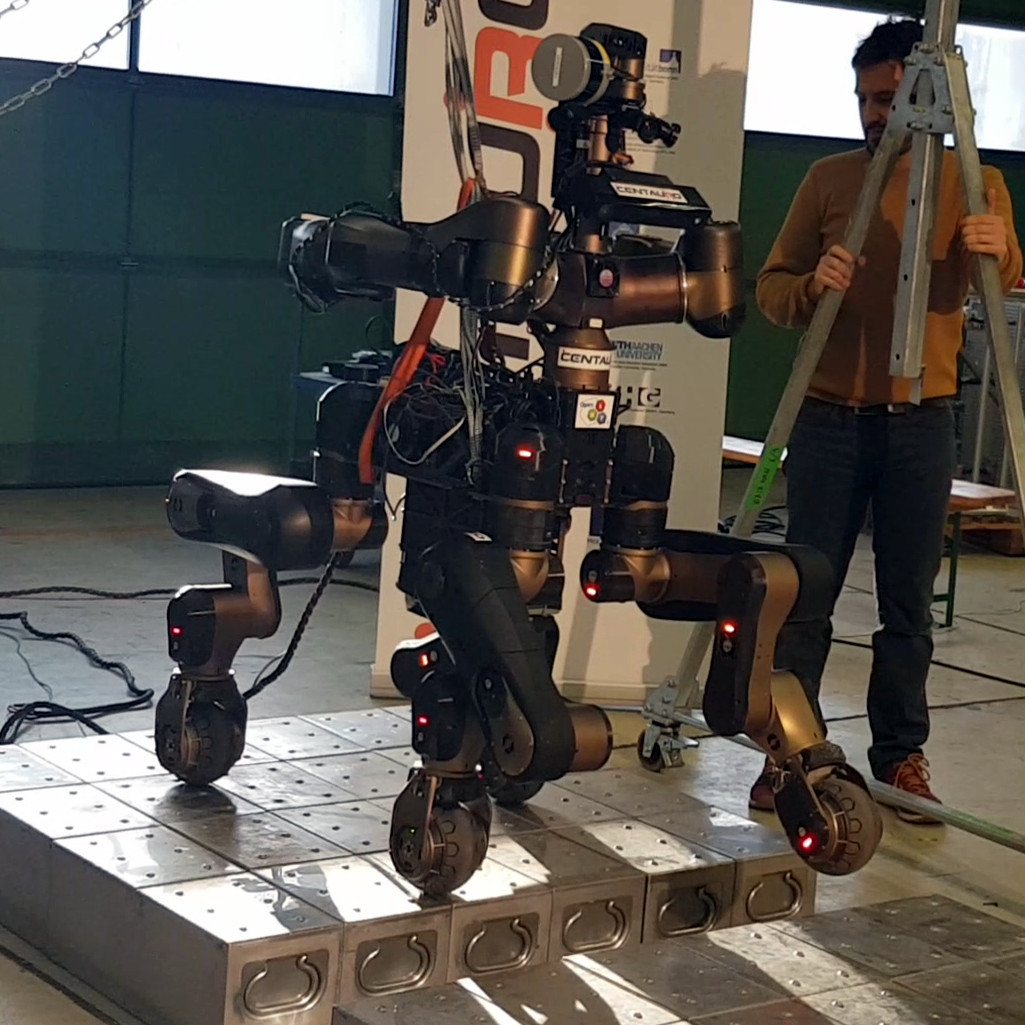}
  \includegraphics[height=\gapheight,clip,trim=60 0 0 0]{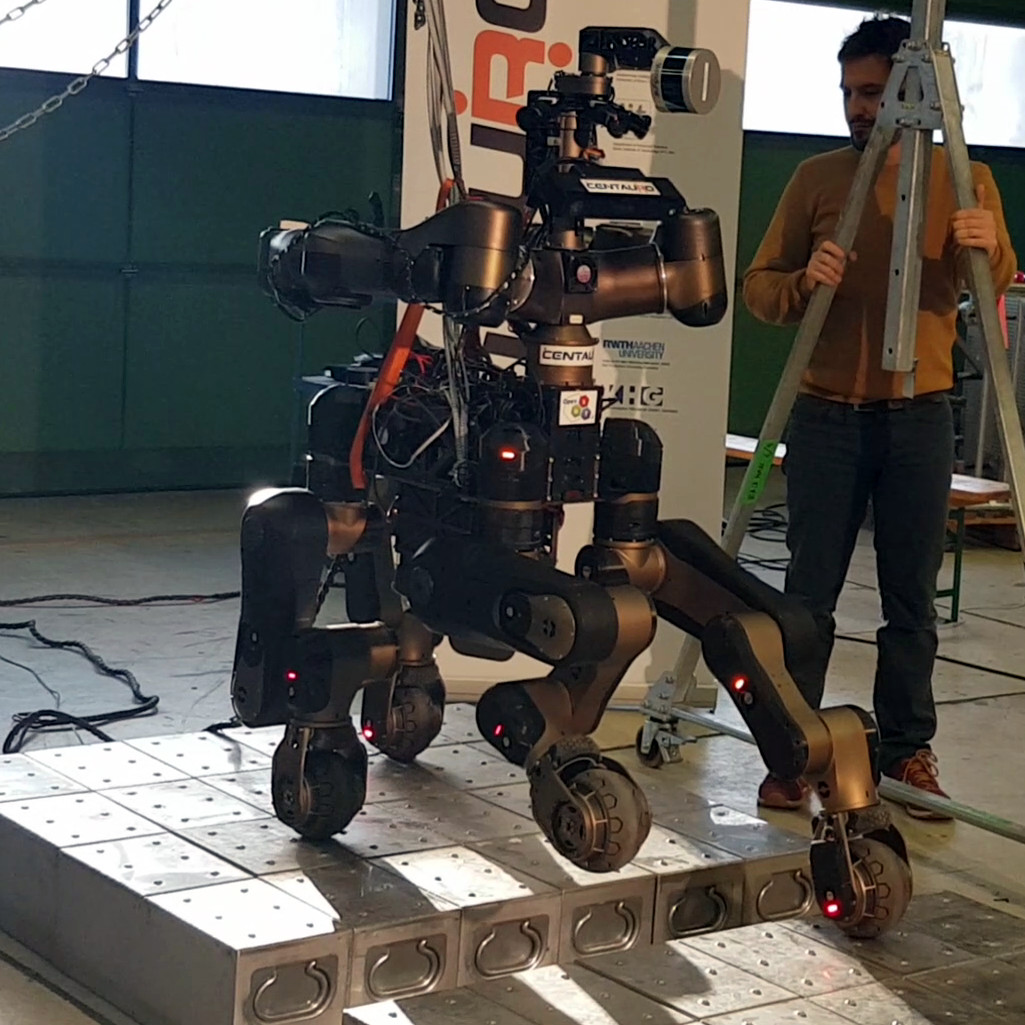}
  \includegraphics[height=\gapheight,clip,trim=60 0 0 0]{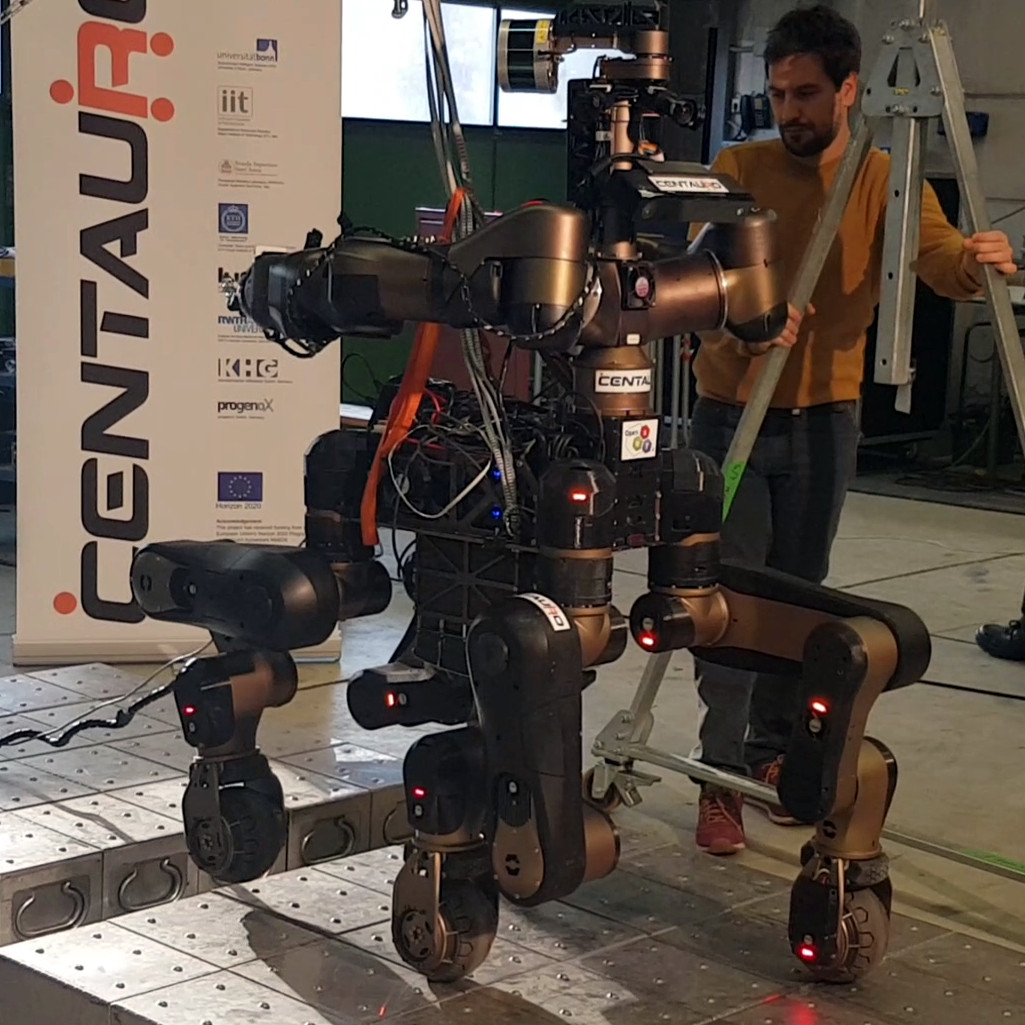}
  \caption{Overcoming a gap with the Centauro robot.}
  \label{fig:eval_gap}
  \vspace{-0.5cm}
\end{figure}

More complex locomotion capabilities were tested in the gap and step field
tests. The gap test required the robot to overcome a 30\,cm gap, which was accomplished
using predesigned stepping motion primitives, which where interleaved with joystick
driving commands (see \cref{fig:eval_gap}).

\begin{figure}[b]
 \centering\newlength{\stairsheight}\setlength{\stairsheight}{2.8cm}
 \includegraphics[height=\stairsheight]{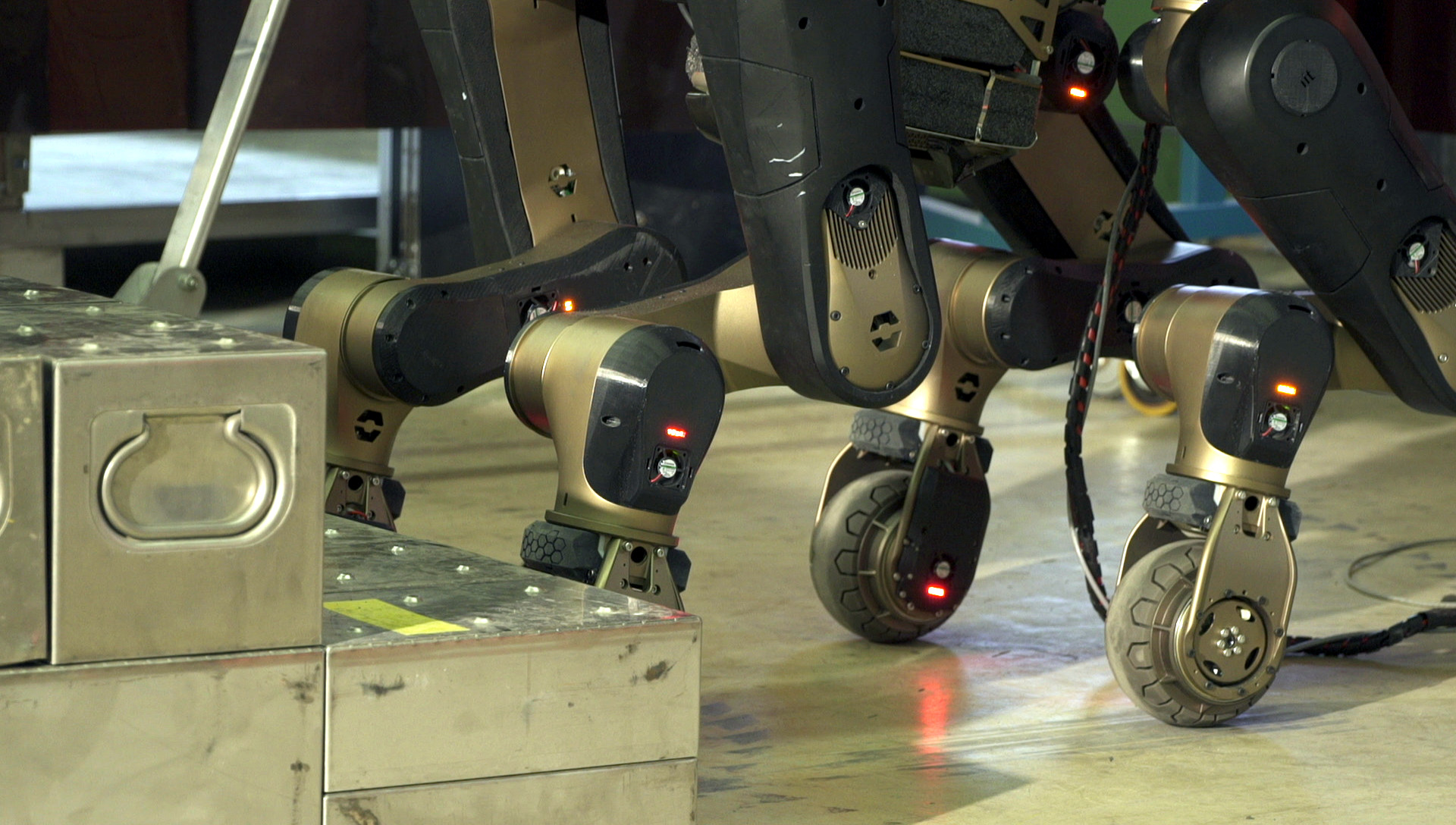}
 \includegraphics[height=\stairsheight]{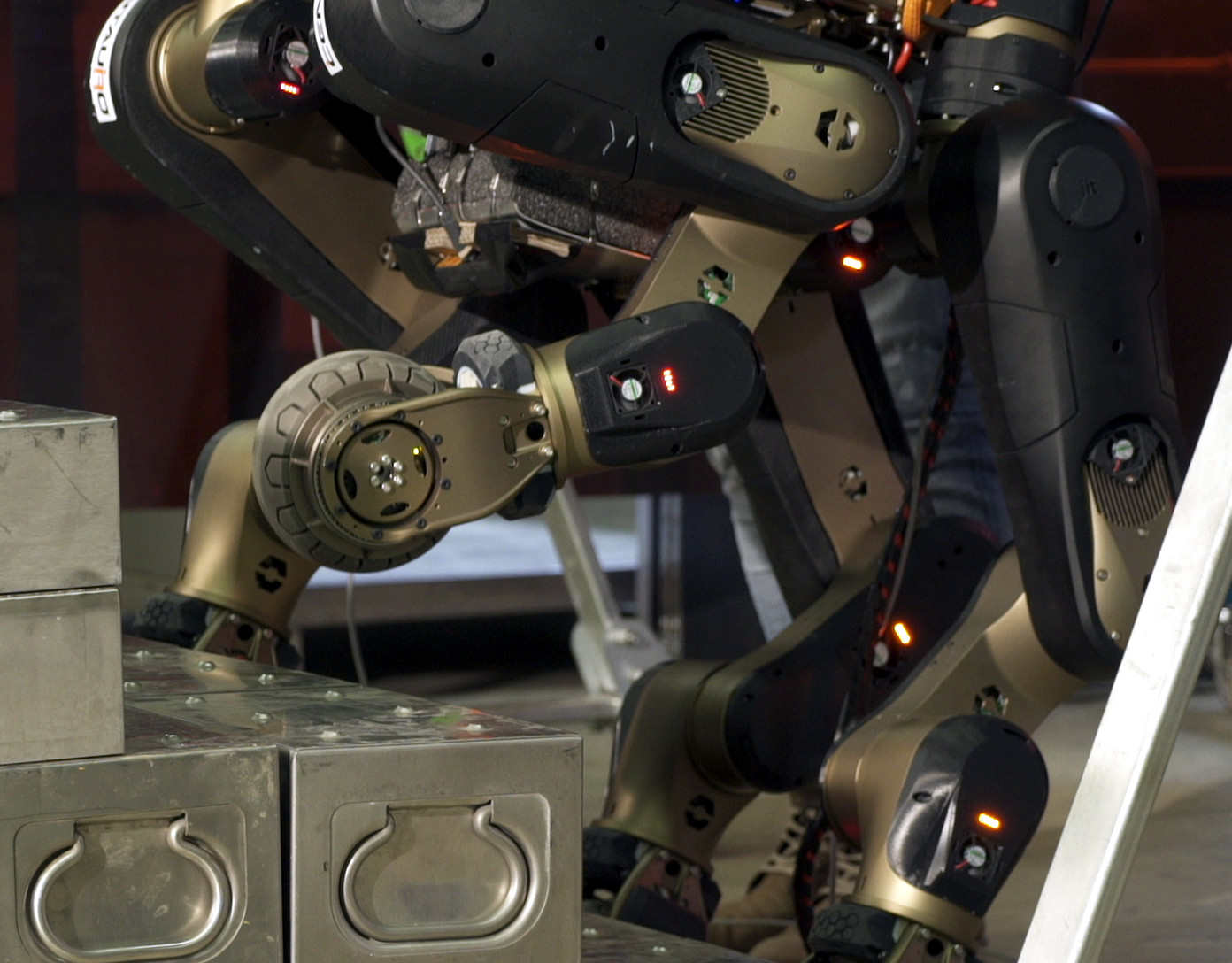}\vspace*{-1ex}
 \caption{Climbing stairs.}
  \vspace{-0.2cm}
 \label{fig:eval_stairs}
\end{figure}

A more challenging test was performed by climbing a set of stairs (see \cref{fig:eval_stairs}).
For this purpose, motion primitives were designed offline before the test,
and executed under supervision of the operators, who could take corrective
actions using the joystick input.
Due to hardware problems, it was only possible to make one serious attempt at
climbing the stairs, which had to be stopped after an actuator shutdown halfway
up---with the robot at least completely on the stairs.

\begin{figure}[b]
 \centering\newlength{\stepheight}\setlength{\stepheight}{2.8cm}
 \includegraphics[height=\stepheight]{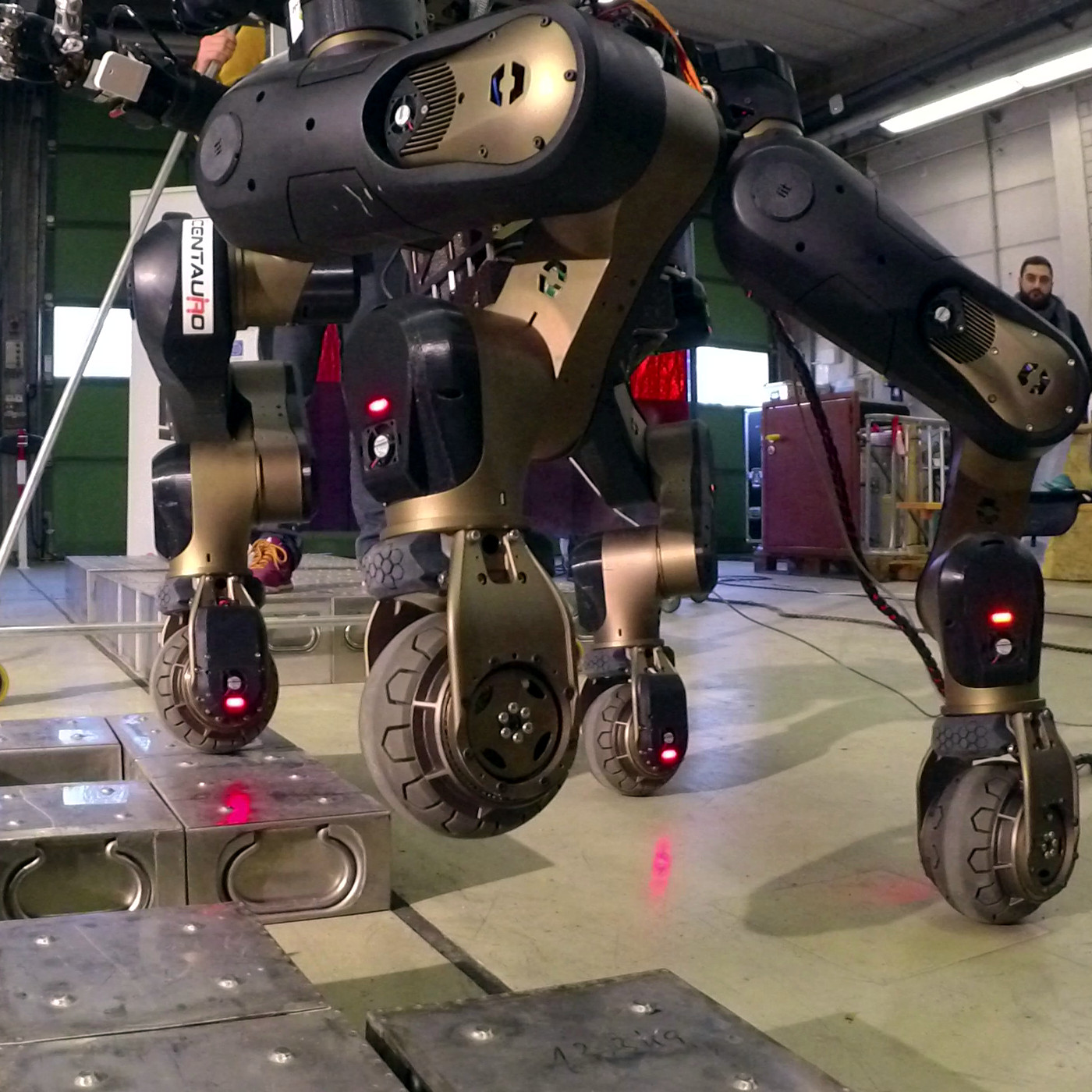}\hfill
 \includegraphics[height=\stepheight]{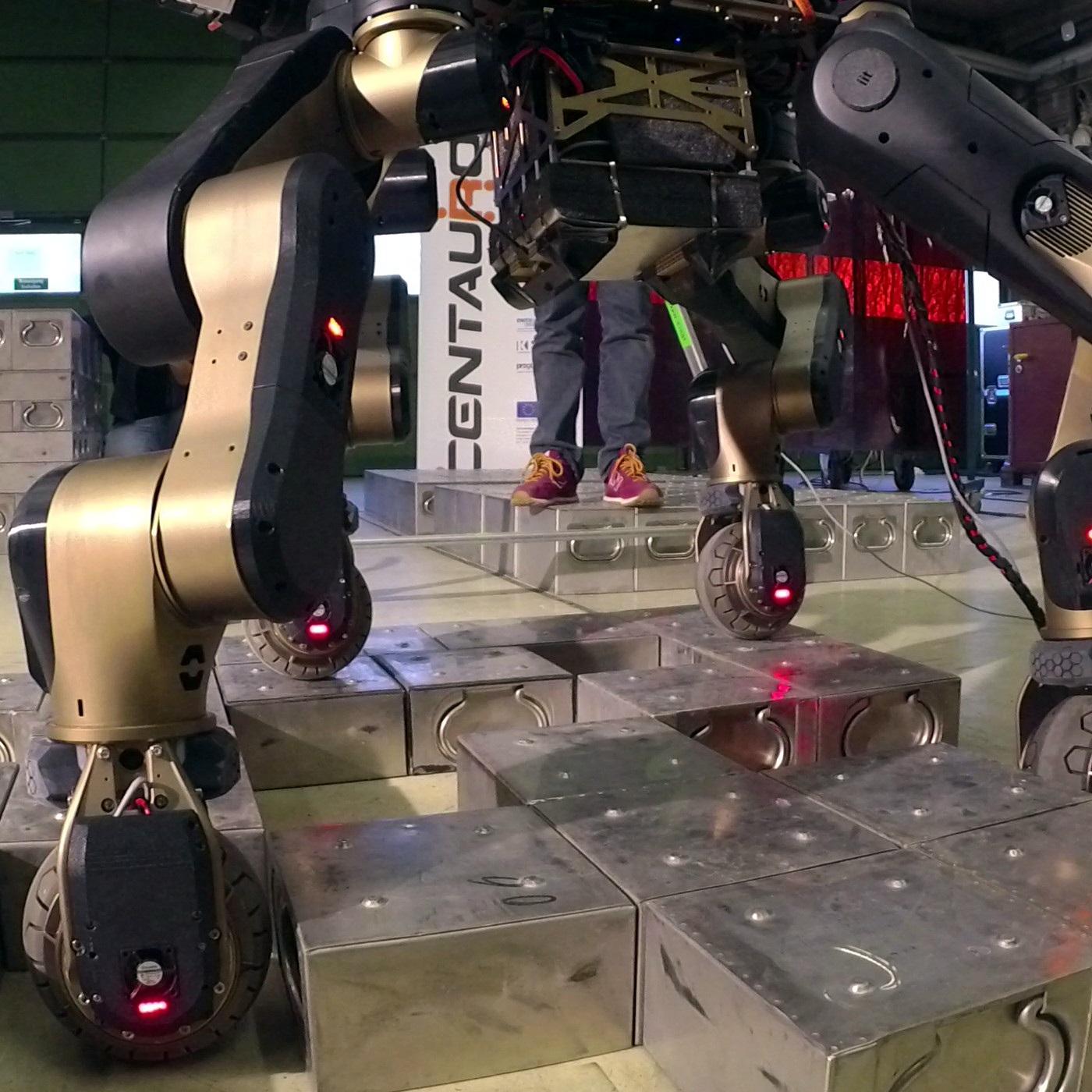}\hfill
 \includegraphics[height=\stepheight]{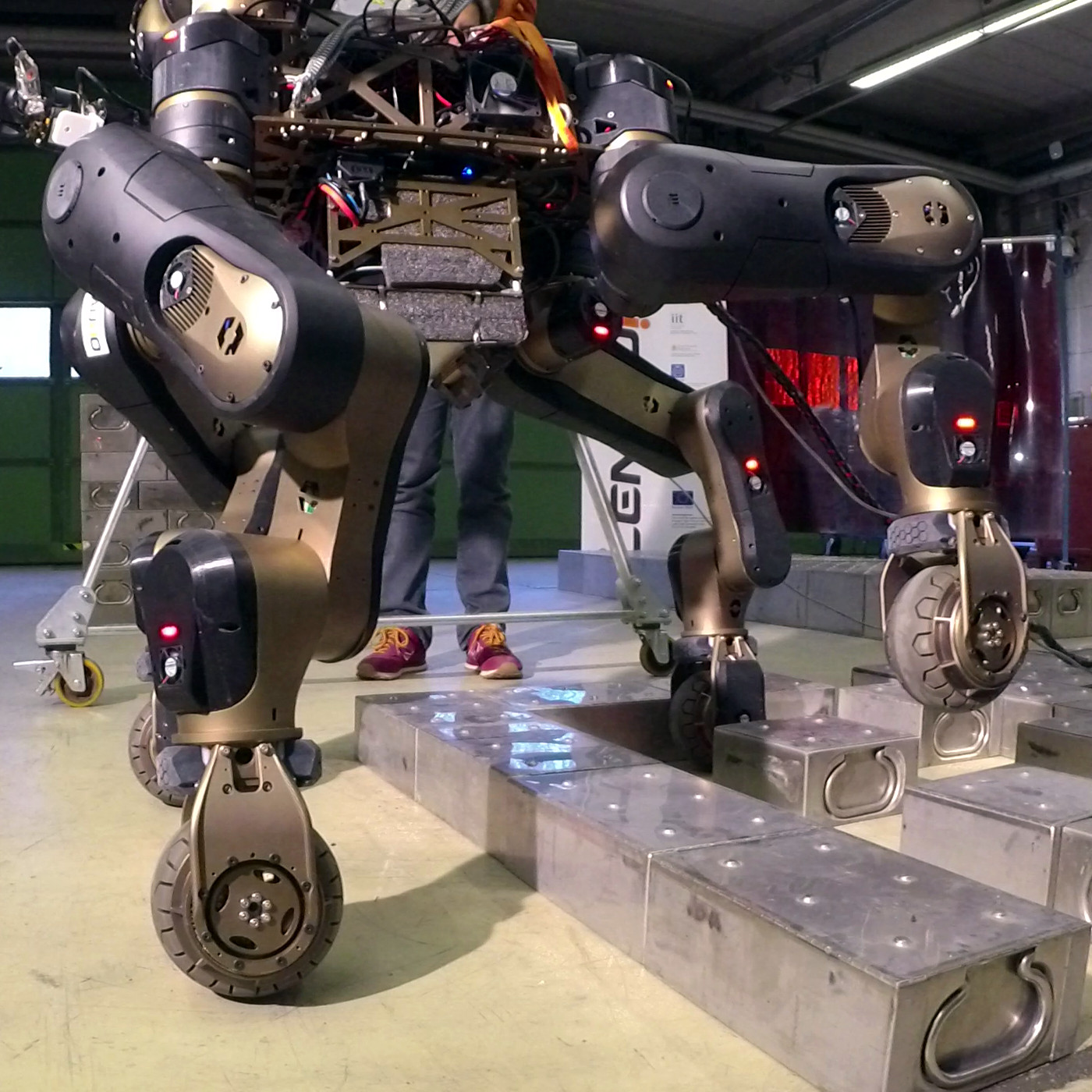}\vspace*{-1ex}\hfill
 \caption{Traversing a step field with the Centauro robot.}
  \label{fig:eval_step}
\end{figure}

Another task was to traverse a step field consisting of 20$\times$20$\times$10\,cm blocks which were placed
on the ground (see \cref{fig:eval_step}). The operators issued stepping commands via the
semi-automatic stepping GUI described in Sec.\,\ref{sec:semi_autonomous_stepping}.
The task was solved two out of two attempts.

Overall, the locomotion capabilities were demonstrated successfully during the Karlsruhe evaluation.
The more complex tasks would have been impossible to finish in acceptable time without autonomy functions.

\subsection{Telemanipulation Tasks}

The first task required the robot to sweep a planar surface with a (dummy)
radiation sensor without touching the surface.
This task was successfully performed using the 6D mouse for wrist control and locomotion via joystick.

\begin{figure}
 \centering\newlength{\plugheight}\setlength{\plugheight}{2.6cm}
 \includegraphics[height=\plugheight]{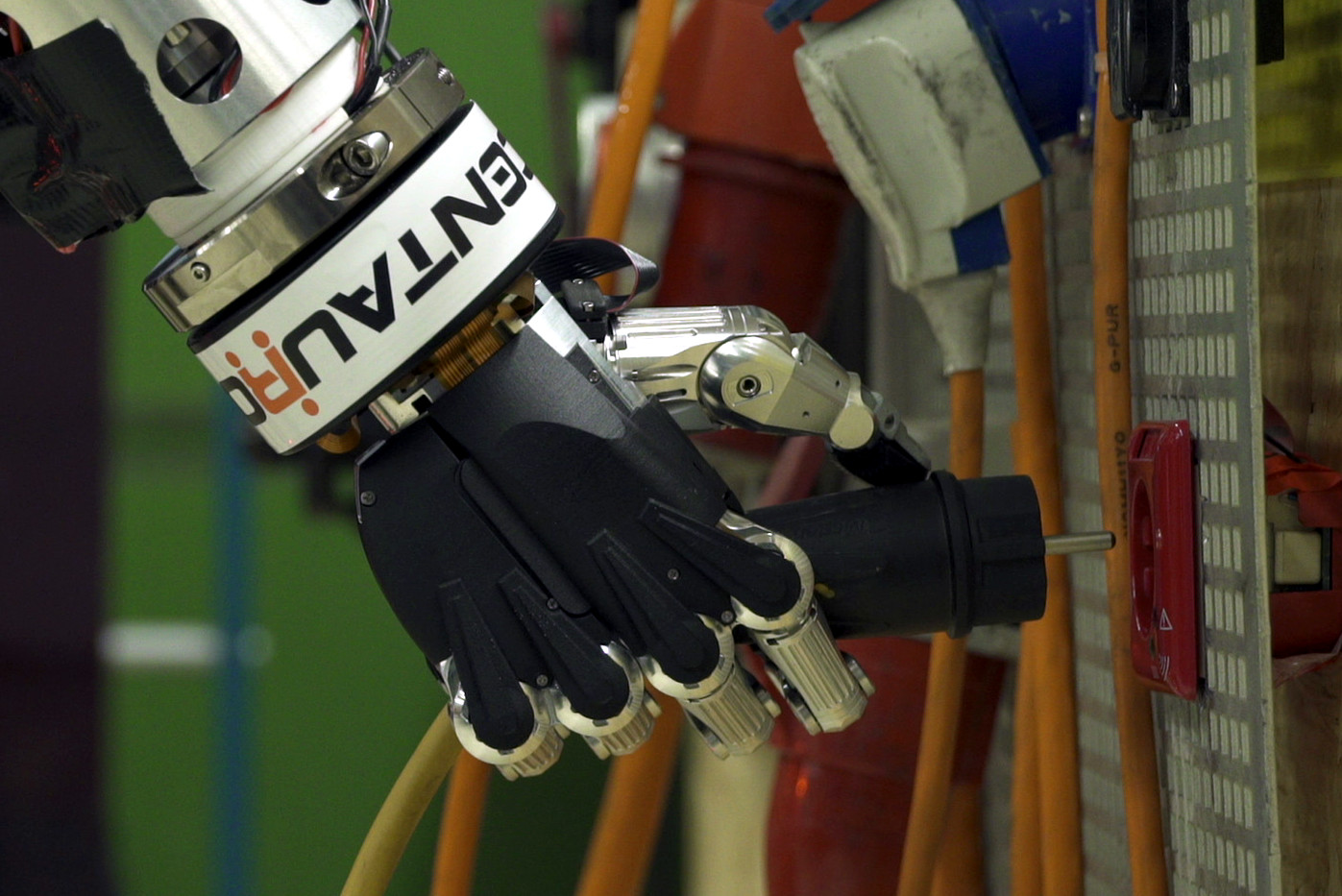}\hfill
 \includegraphics[height=\plugheight]{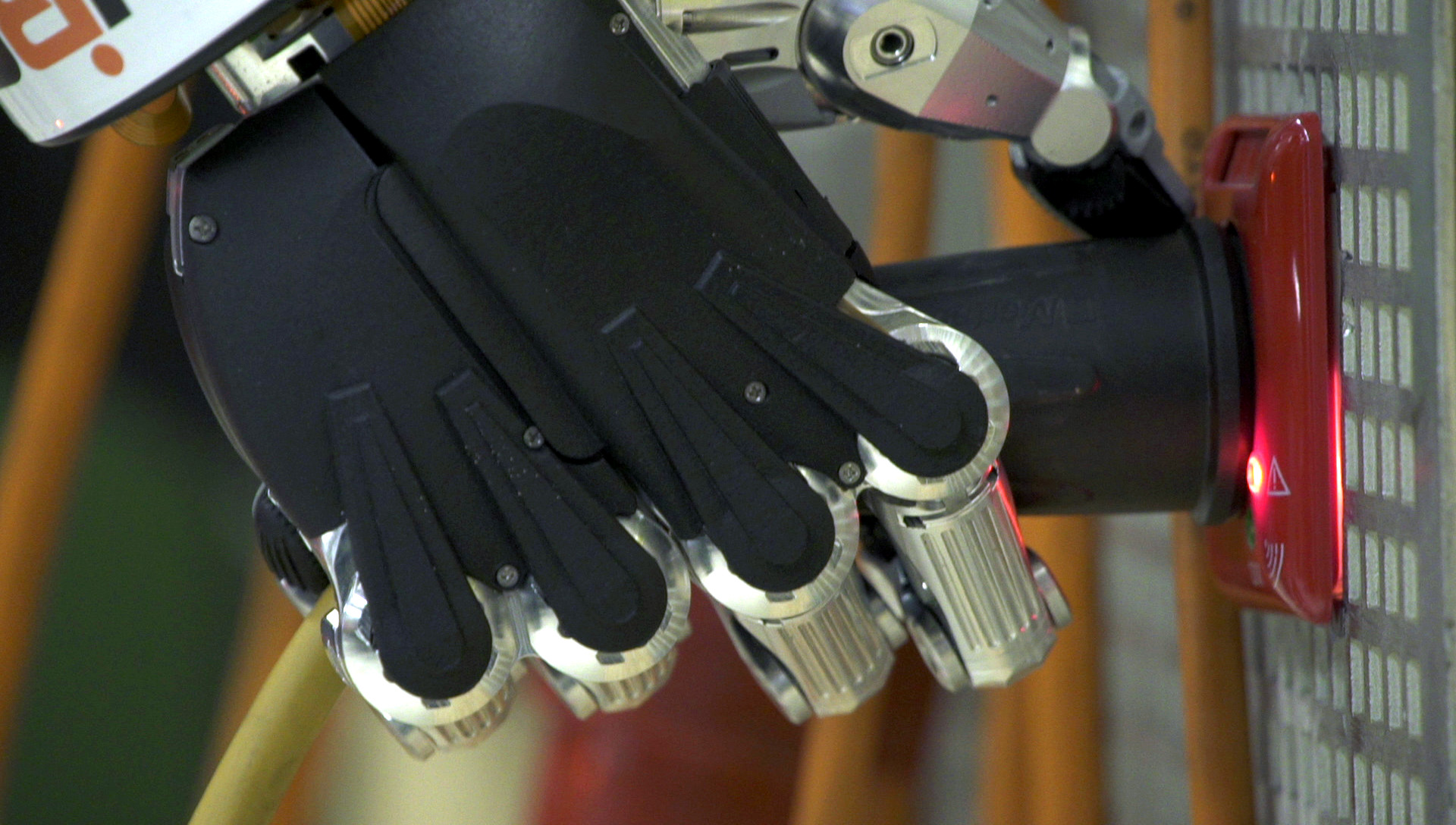}\vspace*{-1ex}
 \caption{Inserting an electrical plug.}
 \label{fig:eval_plug}
\end{figure}

An electrical plug had to be inserted by the
robot (\cref{fig:eval_plug}), which was performed using the 6D mouse.
After two successful attempts, a plastic part in the robot wrist
broke due to excessive force during the third attempt---the operators
had misjudged the situation slightly.

\begin{figure}
 \centering
 \includegraphics[height=2.6cm,clip,trim=325 0 25 100]{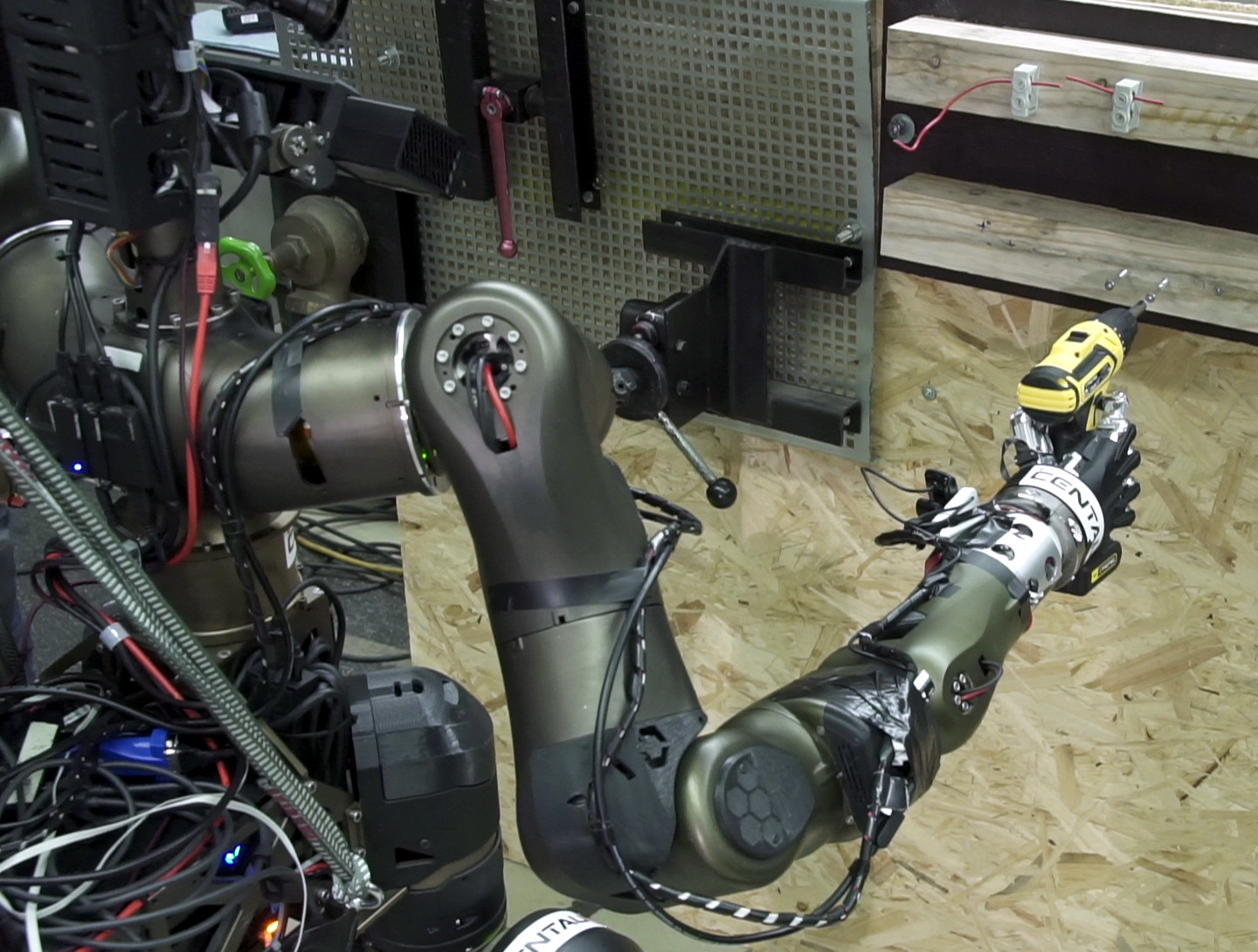}\hfill
 \includegraphics[height=2.6cm,clip,trim=0 0 100 0]{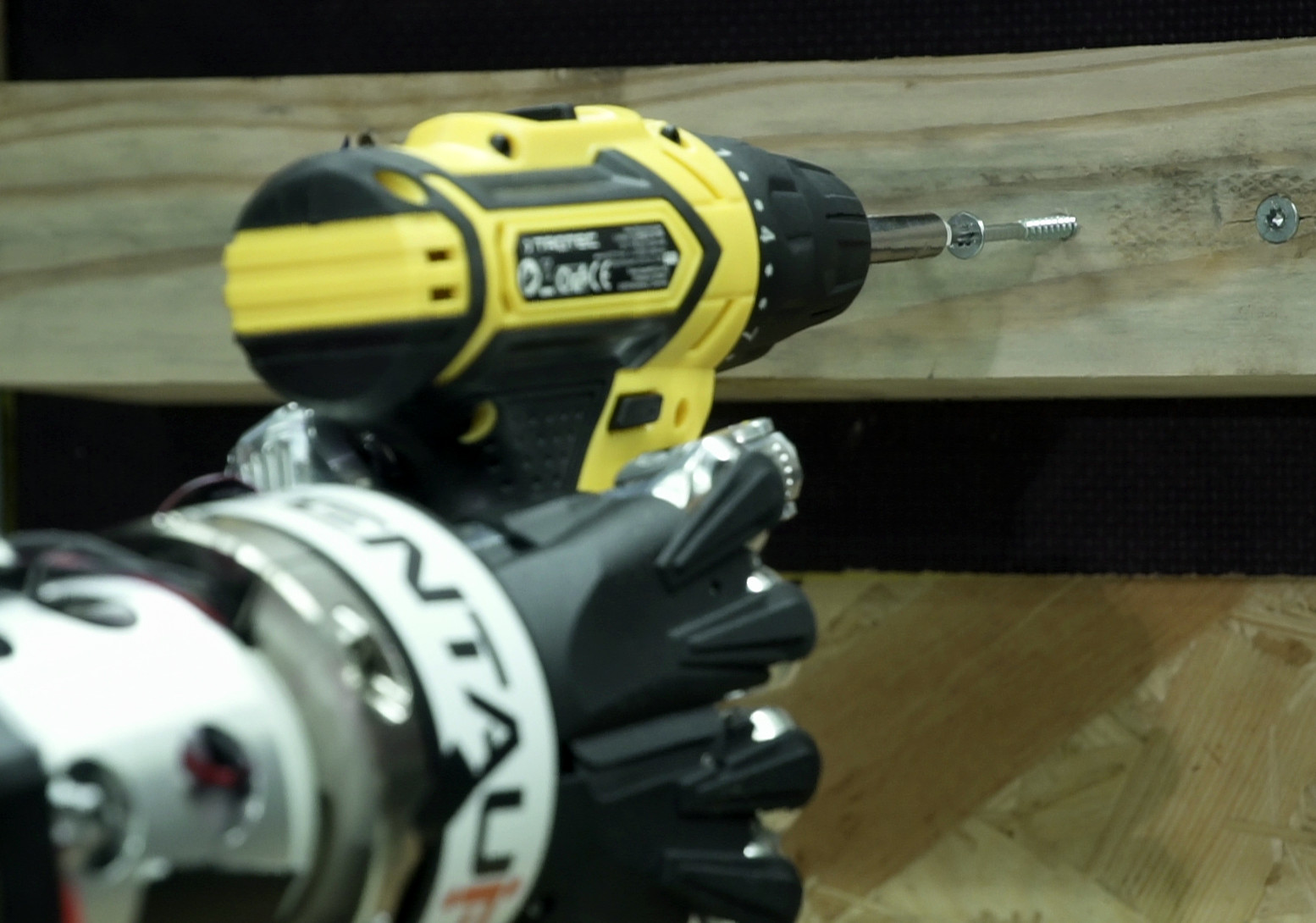}\hfill
 \includegraphics[height=2.6cm,clip,trim=0 0 100 0]{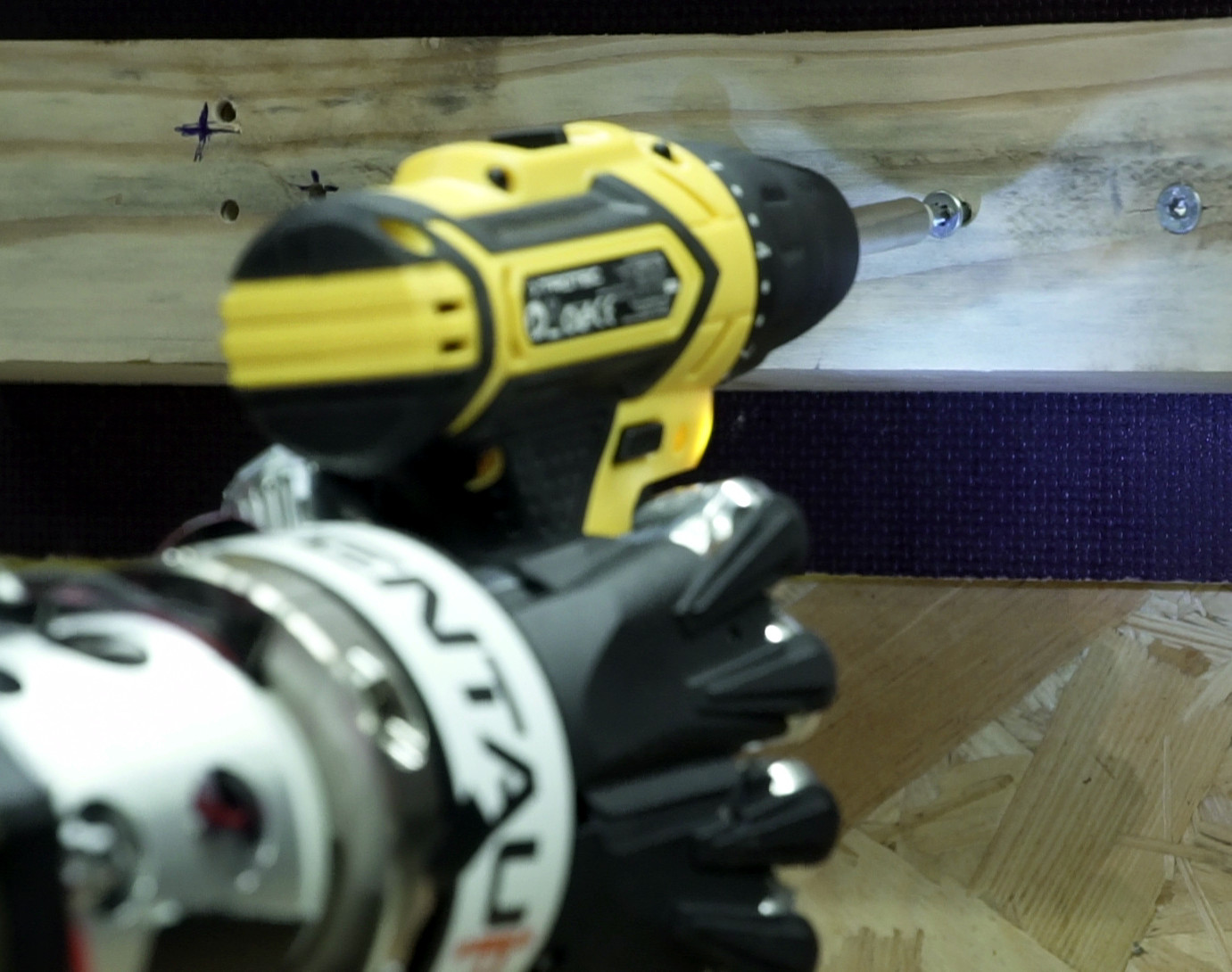}\vspace*{-1ex}
 \caption{Driving a screw into plywood.
 Left: Robot arm in front of the screw.
 Center/right: Detail on fine alignment and screwing.}
 \label{fig:eval_screw}

\end{figure}

The most complex telemanipulation task required the robot to drive a screw into a wooden block (\cref{fig:eval_screw}).
The robot used a cordless screw driver for this task, starting with the tool in hand.
The wooden block was approached using joystick locomotion,
mainly guided by camera images and the 3D laser scanner point cloud.
Next, the tip of the screw driver was aligned with the
screw using 6D mouse control, guided by camera images.
For gaining an additional perspective, a small webcam was mounted on the other hand,
providing a controllable-viewpoint perspective to the operators.
After alignment was visually confirmed, the cordless screwdriver was activated using
the index finger of the robot. During the screwing process, the operators had
to ensure that the tool tip was in constant contact with the screw head, which was facilitated using the single-axis
mode of the 6D mouse interface.
Overall, three out of three attempts were successful.

\subsection{Autonomous Manipulation}

\begin{figure}[t]
 \centering\newlength{\graspheight}\setlength{\graspheight}{2.85cm}
 \includegraphics[height=\graspheight]{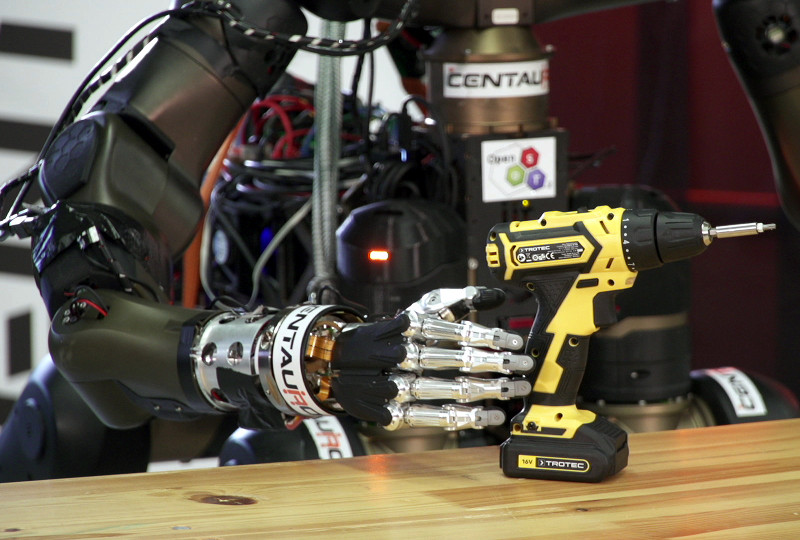}\hfill
 \includegraphics[height=\graspheight]{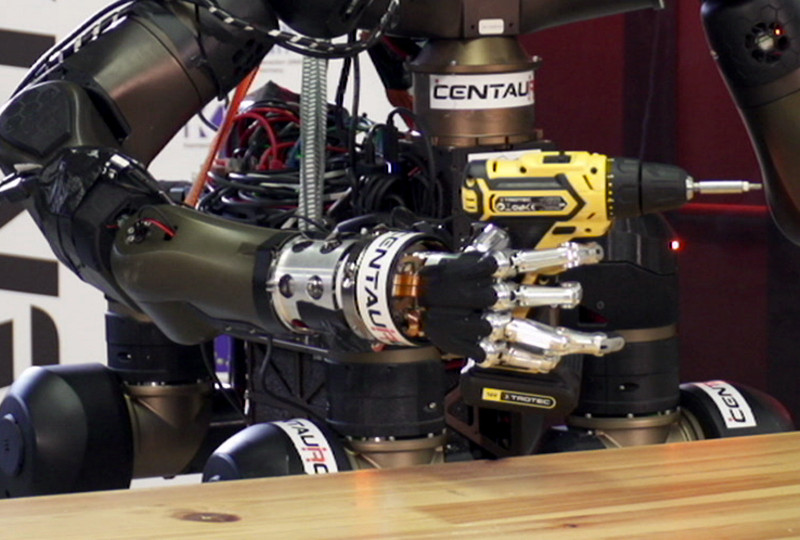}\vspace*{-1ex}
 \caption{Autonomous Grasping: approaching (l.) and lifting the drill (r.).}
 \label{fig:eval_grasping}
 \vspace{-0.5cm}
\end{figure}

The objective of this test was to detect, segment, and estimate the pose of a previously
unknown cordless driller in front of the robot (\cref{fig:eval_grasping}). 
After pose estimation, a grasping pose was to be transferred from a known model to
the new instance and the driller was to be grasped.

We performed this experiment many times, since it had a higher failure
rate due to the complexity and the number of involved components.
While the system performed well on the operator side, failures cases on the system side include imprecise segmentation or misregistration,
both resulting in missed grasps, and hardware failures.
Overall, the success rate improved during testing.

\section{Conclusion}

On the example of the Centauro robot, we successfully demonstrated several useful
autonomous functions that assist the operators on different levels of autonomy.
Their efficiency was especially demonstrated considering that all experiments were performed without any previous training.
Operation time was often shortened or task fulfillment was enabled.
We are convinced that such strong autonomy functions are needed for disaster response
robots to make rapid deployment in unknown scenarios possible.



\bibliographystyle{IEEEtranN}
\bibliography{references}

\addtolength{\textheight}{-12cm}   





\end{document}

%% file: figures/figures_section1/centauro_overview.pgf
\begin{tikzpicture}[
 	font=\sffamily\footnotesize,
    every node/.append style={text depth=.2ex},
	box/.style={rectangle, inner sep=0.5, anchor=west},
	line/.style={red, thick},
	l/.style={font=\sffamily\scriptsize},
]


\node[anchor=south west,inner sep=0] (image) at (0,0) {\includegraphics[width=0.6\linewidth]{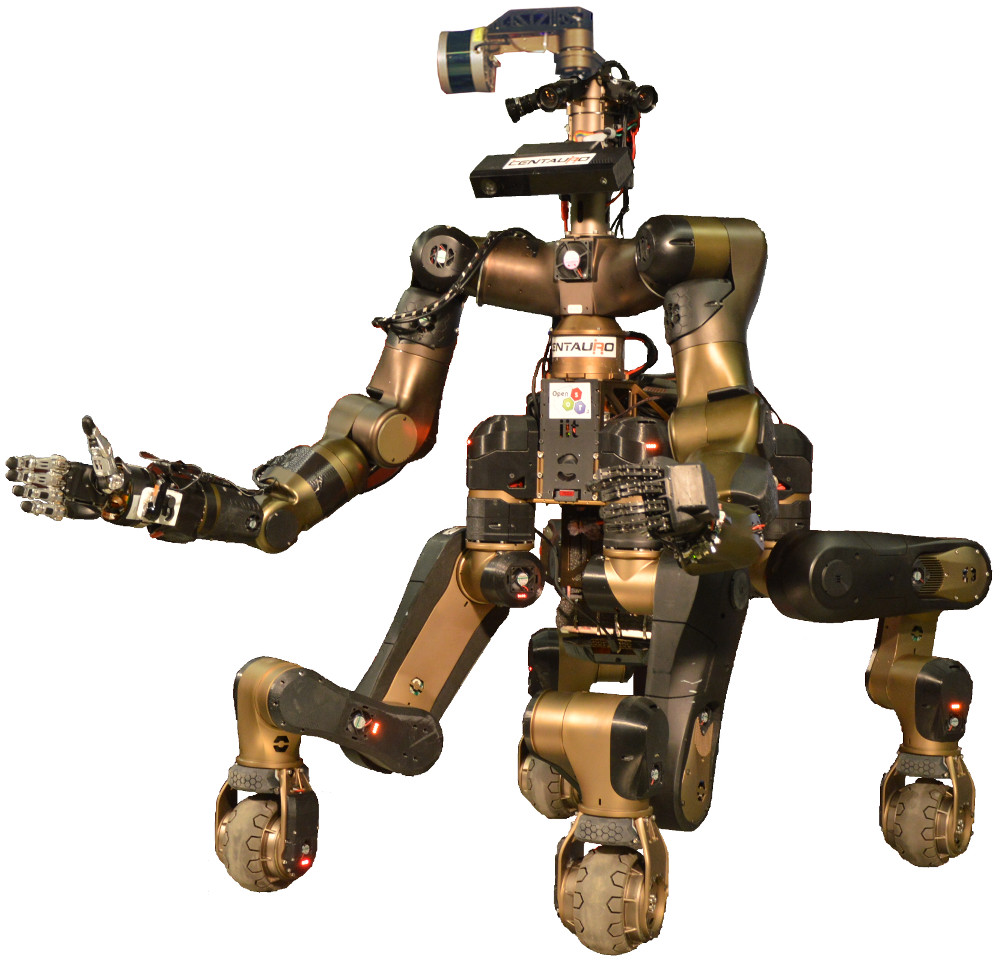}};

\node[box, align=center,l](laser_scanner) at(-1.4,5.1){3D laser\\scanner};
\draw[line](laser_scanner.south west)--(laser_scanner.south east);
\draw[line](laser_scanner.south east)--(2.3,4.8);

\node[box,l](cameras) at(5.3 ,5.2){Cameras};
\coordinate(camera_split) at (3.3,5);
\draw[line](cameras.south west)--(cameras.south east);
\draw[line](cameras.south west)--(camera_split);
\draw[line](camera_split)--(3.3, 4.6);
\draw[line](camera_split)--(2.7,4.5);
\draw[line](camera_split)--(3.0,4.6);

\node[box,align=center,l](kinect) at(-1.4,4.2){RGB-D\\sensor};
\draw[line](kinect.south west)--(kinect.south east);
\draw[line](kinect.south east)--(2.5,4.15);

\node[box,l](arm) at(-1.4,3.3){7 DoF arm};
\draw[line](arm.south west)--(arm.south east);
\draw[line](arm.south east)--(1.5,2.6);

\node[box,align=center,l](schunk) at(-1.4,1.8){9 DoF\\Schunk hand};
\draw[line](schunk.south west)--(schunk.south east);
\draw[line](schunk.south east)--(0.7,2.5);

\node[box,align=center,l](soft_hand) at(5.05,3){1 DoF\\flexible hand};
\draw[line](soft_hand.south west)--(soft_hand.south east);
\draw[line](soft_hand.south west)--(3.6,2.5);

\node[box,align=center,l](leg) at(5.8,1.0){5 DoF\\leg};
\draw[line](leg.south west)--(leg.south east);
\draw[line](leg.south west)--(4.9,1.5);

\node[box,align=center,l](wheel) at(-1.4,0.6){360\textdegree\, steerable\\wheel};
\draw[line](wheel.south west)--(wheel.south east);
\draw[line](wheel.south east)--(1.35,0.75);

\node[box, align=left,l](base) at(5.1,4.2){Base with\\CPU, router\\and battery};
\draw[line](base.south west)--(base.south east);
\draw[line](base.south west)--(4.1,2.7);

%
%

\end{tikzpicture}

%% file: figures/teleoperation_architecture/situation_awareness.pgf
\begin{tikzpicture}[
 	font=\sffamily\footnotesize,
    every node/.append style={text depth=.2ex},
	box/.style={rectangle, inner sep=0.5, anchor=west},
	line/.style={red, thick},
 	l/.style={font=\sffamily\scriptsize},
]



\draw (0.0,0.0) rectangle ++(5, 2.8);
\draw (5.05,0.0) rectangle ++(5, 2.8);
\draw (10.1,0.0) rectangle ++(5, 2.8);

\node[anchor=south west,inner sep=0] (image) at (0.1,0.5) {\includegraphics[width=2.2cm]{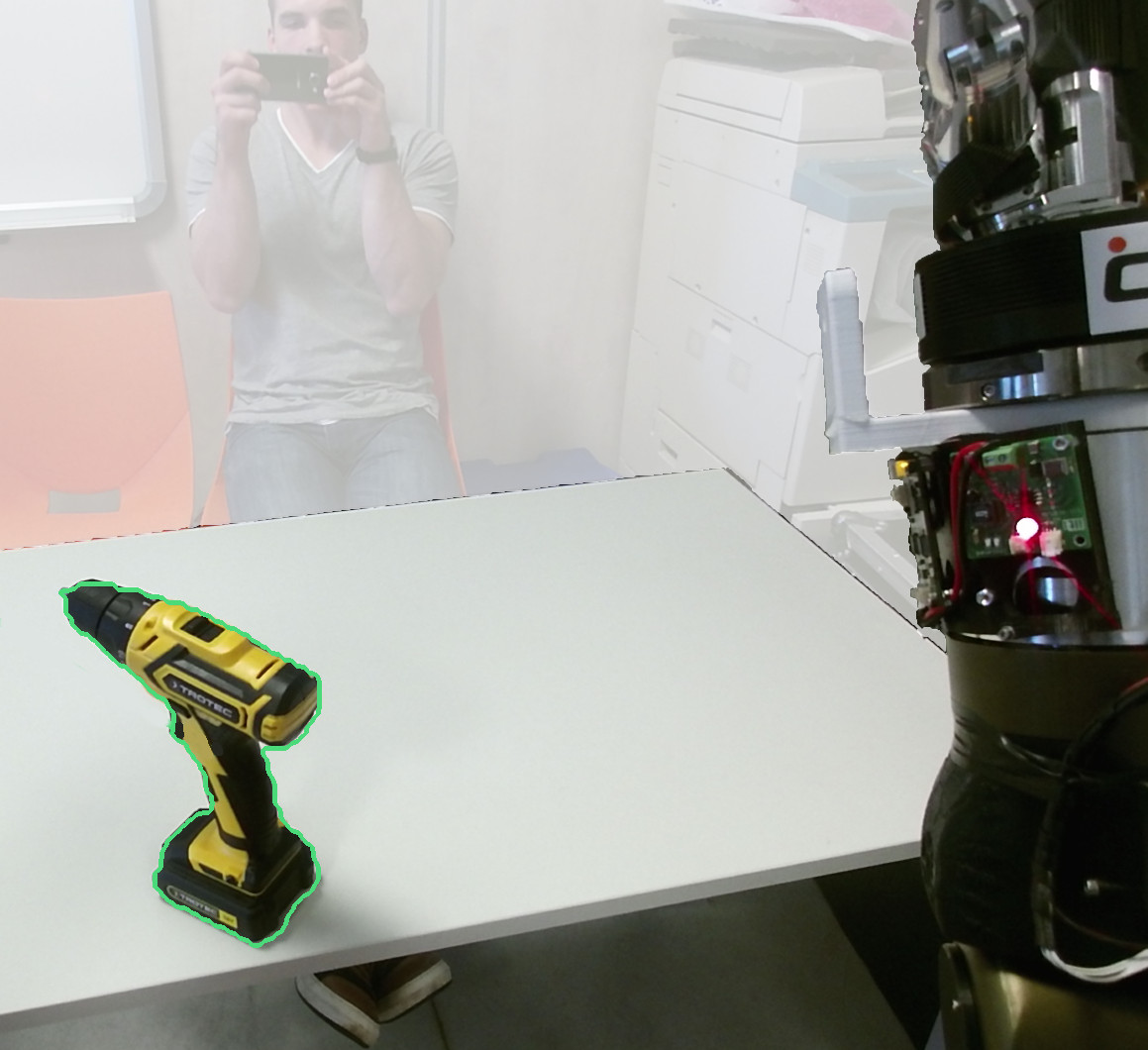}};
\node[l,box, align=center](laser_scanner) at(0.1,0.25){Drill 0.969};
\draw[green](0.35,0.4) -- ++(0.1,0.45);

\node[anchor=south west,inner sep=0] (image) at (2.4,0.5) {\includegraphics[width=2.5cm,height=2cm]{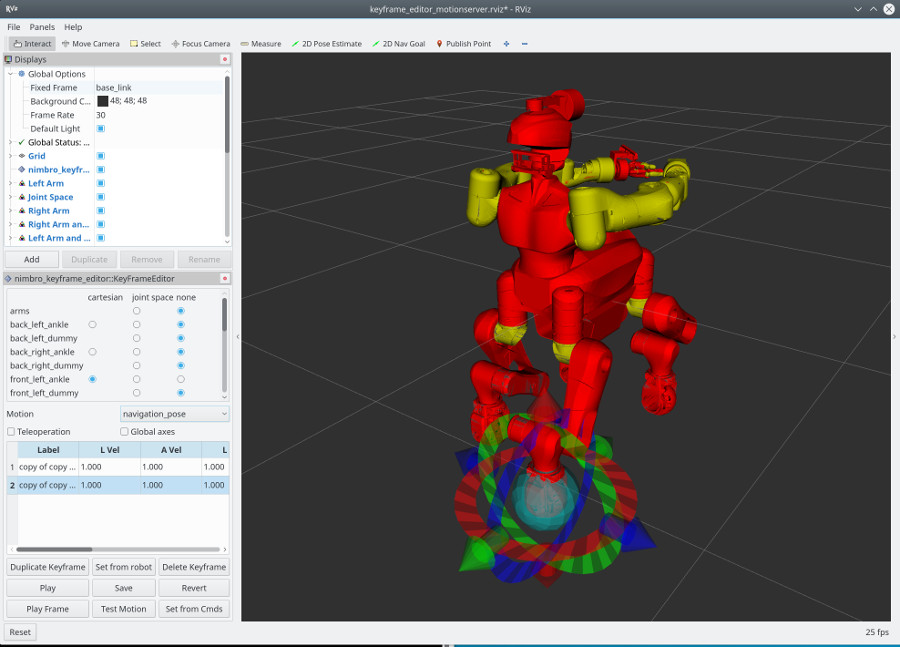}};

\node[anchor=south west,inner sep=0] (image) at (5.1,0.02) {\includegraphics[width=4.9cm,height=2.7cm]{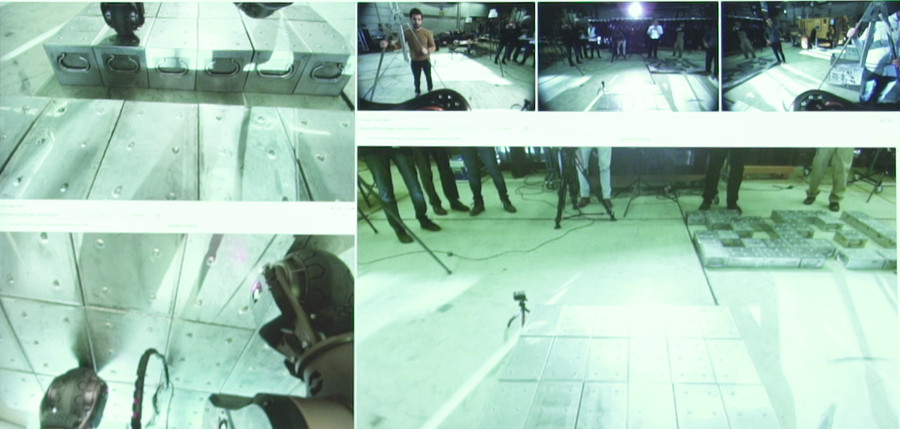}};

\node[anchor=south west,inner sep=0] (image) at (12.2,0.02) {\includegraphics[width=2.8cm,height=2.7cm]{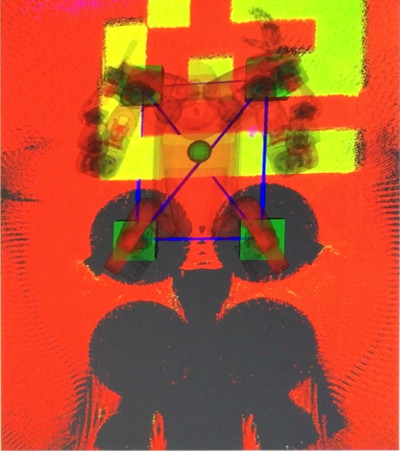}};

\node[anchor=south west,inner sep=0] (image) at (10.15,0.7) {\includegraphics[width=2cm,height=1.5cm]{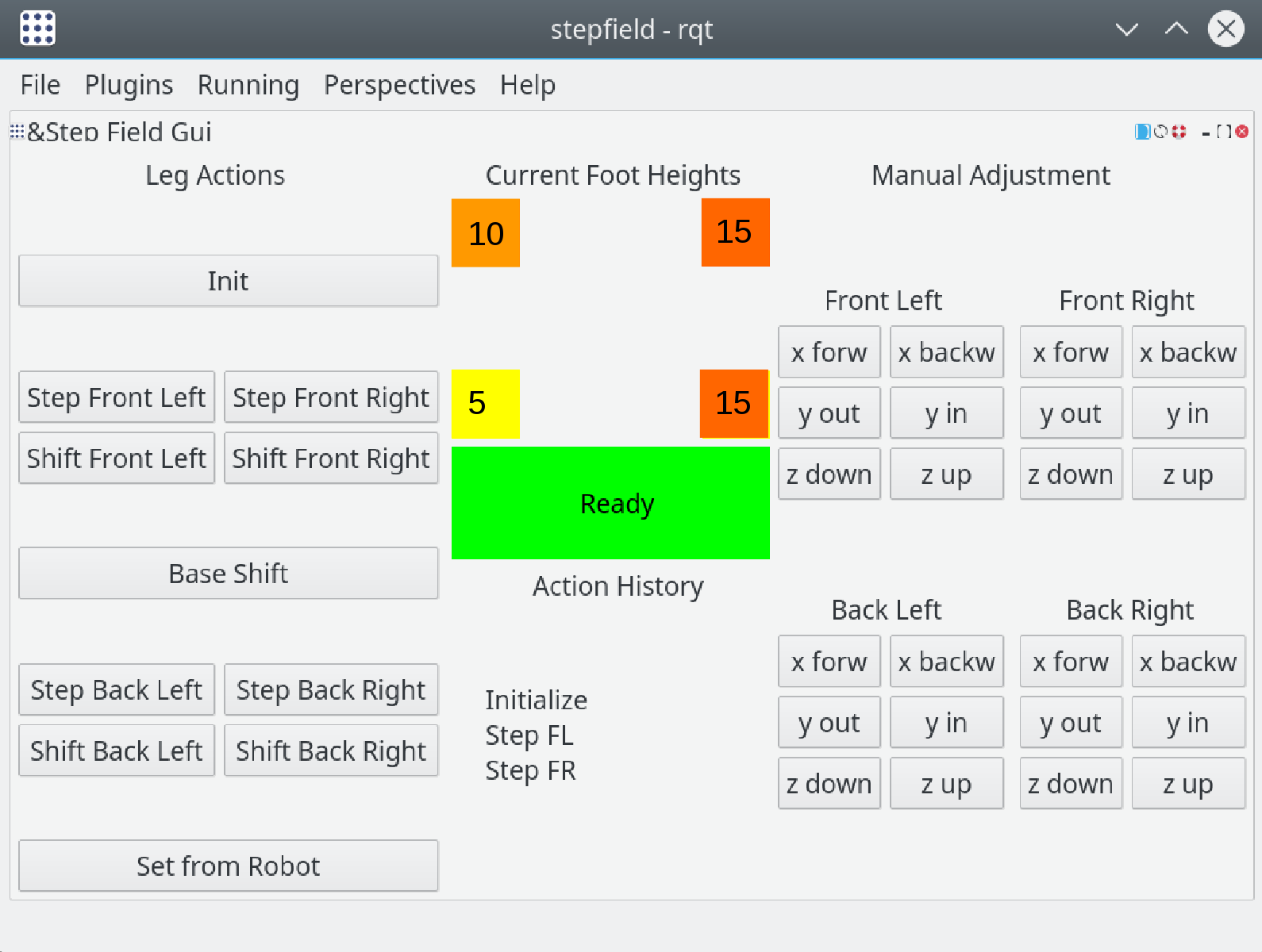}};


\node[l,box, align=center, color=red](laser_scanner) at(0.35,3.2){RGB-D image\\with annotation};
\draw[red,dashed](0.35,2.95) -- (0.1,2.55);
\draw[red,dashed](2.05,2.95) -- (2.3,2.55);
\node[l,box, align=center, color=red](laser_scanner) at(2.77,3.2){Robot state \&\\Keyframe editor};
\draw[red,dashed](2.77,2.95) -- (2.4,2.55);
\draw[red,dashed](4.53,2.95) -- (4.9,2.55);

\node[l,box, align=center, color=red](laser_scanner) at(7.53,3.2){Panoramic view \&\\RGB Kinect image};
\draw[red,dashed](7.5,2.95) -- (7.05,2.75);
\draw[red,dashed](9.6,2.95) -- (10.05,2.75);

\node[l,box, align=center, color=red](laser_scanner) at(5.6,3.2){Foot\\cameras};
\draw[red,dashed](5.6,2.95) -- (5.1,2.75);
\draw[red,dashed](6.58,2.95) -- (7,2.75);

\node[l,box, align=center, color=red](laser_scanner) at(12.3,3.2){Pointcloud, ground\\contact \& COM markers};
\draw[red,dashed](12.3,2.95) -- (12.2,2.75);
\draw[red,dashed](14.97,2.95) -- (15.05,2.75);

\node[l,box, align=center, color=red](laser_scanner) at(10.4,3.2){Task specific\\GUI };
\draw[red,dashed](10.35,2.95) -- (10.15,2.25);
\draw[red,dashed](12.0,2.95) -- (12.15,2.25);

\node[l,box, align=center](laser_scanner) at(2.1,-0.25){Monitor 1};
\node[l,box, align=center](laser_scanner) at(7.1,-0.25){Monitor 2};
\node[l,box, align=center](laser_scanner) at(12.1,-0.25){Monitor 3};

\end{tikzpicture}

%% file: figures/locomotion_control/locomotion_control.pgf
\begin{tikzpicture}[
 	font=\sffamily\footnotesize,
    every node/.append style={text depth=.2ex},
	box/.style={rectangle, inner sep=0.5, anchor=west},
	line/.style={red, thick},
 	l/.style={font=\sffamily\scriptsize},
]


\node[anchor=south west,inner sep=0] (image) at (0.0,0.5) {\includegraphics[height=3.05cm]{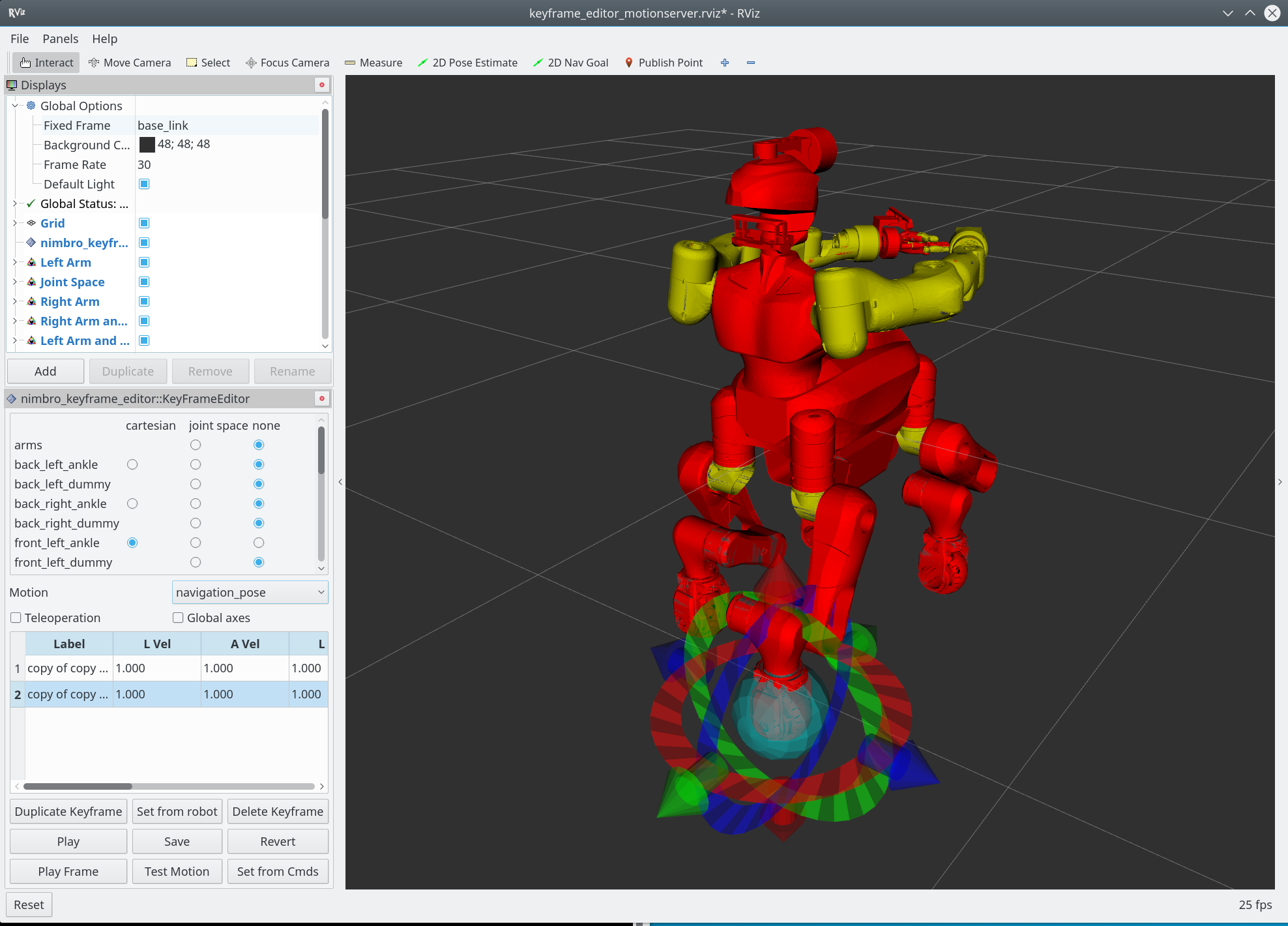}};
\node[anchor=south west,inner sep=0] (image) at (4.4,0.5) {\includegraphics[height=3.05cm]{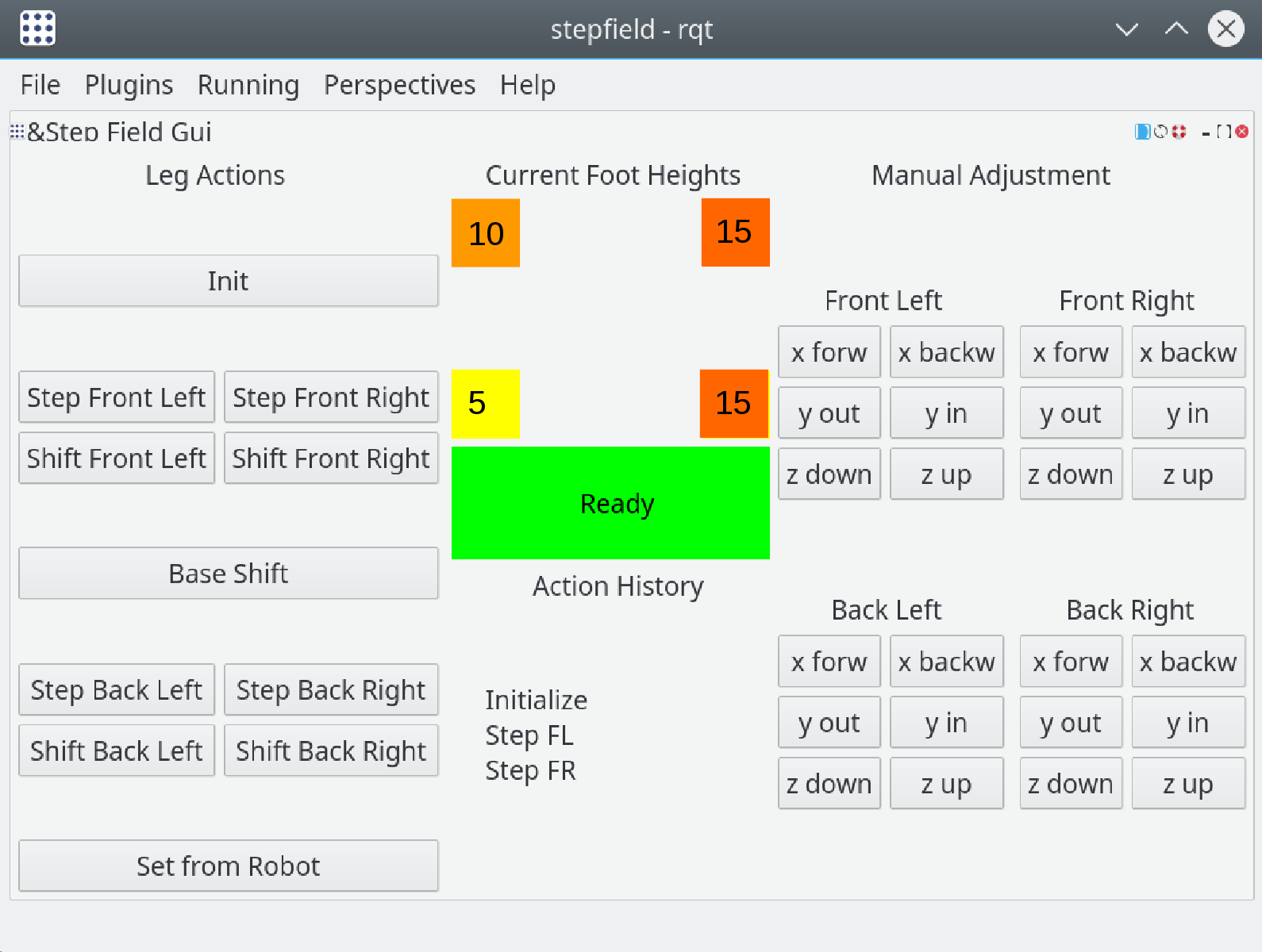}};

\end{tikzpicture}

%% file: figures/manipulation/overview.pgf
\begin{tikzpicture}[
    font=\sffamily\footnotesize,
    every node/.append style={text depth=.2ex},
	sensor/.style={rectangle,rounded corners,draw=black,fill=red!20,align=center},
    module/.style={rectangle,rounded corners,draw=black,fill=yellow!20,align=center},
    perception/.style={rectangle,rounded corners,draw=black,fill=blue!20,align=center},
    group/.style={rectangle,rounded corners,draw=black},
    extmodule/.style={module,dashed,fill=black!10},
    l/.style={font=\sffamily\scriptsize,align=center},
]

\node[sensor] (kinect) {Kinect\\v2};

\node[perception,right=0.5cm of kinect](semsec) {Semantic\\Segmentation};

%
%
\node[perception, right=0.5cm of semsec] (poseest){Pose\\Estimation};

\node[module, right=0.5cm of poseest,yshift=-0.8cm] (graspplan){Grasp\\Planning};

\node[sensor] (laser) at ($(kinect)+(0,-2.cm)$) {3D\\Laser};
\node[perception,right=1.05cm of laser] (slam) {Laser\\SLAM};

\node[module, right=2.8cm of slam] (trajopt){Trajectory\\Optimization};
%
%
%
%
%
\draw[-latex]   (kinect) -- (semsec);
\draw[-latex]   (semsec) -- ++(1,0) |- (poseest);
\draw[-latex]   (semsec) -- ++(1.2,0) |- (graspplan);
\draw[-latex]   (poseest) -| (graspplan);
\draw[-latex]   (graspplan) -- (trajopt);
\draw[-latex]   (laser) -- (slam);
\draw[-latex]   (slam) -- (trajopt);
\draw[-latex]   (trajopt) -- ++(1.5,0);

\node[l, right=0.08cm of poseest,yshift=0.15cm]{6D Pose};
\node[l, right=0.5cm of semsec,yshift=-0.98cm]{Object Cloud};
\node[l, right=0.5cm of slam,yshift=-0.18cm]{Collision Map};
\node[l, below=-0.04cm of graspplan,xshift=-.45cm]{Motion};
\node[l, right=-0.1cm of trajopt,yshift=0.35cm]{Joint\\Trajectory};
%
%
%
%
%
%
%

\end{tikzpicture}